\documentclass[11pt]{article}

\usepackage[preprint]{acl}

\usepackage{times}
\usepackage{latexsym}
\usepackage[most]{tcolorbox} %modify by yz
\usepackage{booktabs}
\usepackage{tabularx}
\usepackage{amsmath}
\usepackage{amssymb}
\definecolor{mydarkgreen}{RGB}{0, 100, 0}
\usepackage{makecell}
\usepackage{xcolor}
\usepackage{hyperref}

\usepackage[table]{xcolor}
\usepackage{colortbl}

\usepackage[T1]{fontenc}
\usepackage[utf8]{inputenc}
\usepackage{enumitem}   % 用于定制列表的缩进
\usepackage{booktabs}   % 用于绘制漂亮的三线表
\usepackage{multirow} 
\usepackage{float}

\usepackage{xcolor}
\usepackage{mdframed}
\usepackage{booktabs}
\usepackage{enumitem}   % 控制 itemize 间距
\usepackage{tabularx}

\usepackage{microtype}

\usepackage{inconsolata}

\usepackage{graphicx}

\definecolor{amberlight}{RGB}{250,238,218}
\definecolor{amberdark}{RGB}{99,56,6}
\definecolor{amberaccent}{RGB}{250,199,117}

\definecolor{bluelight}{RGB}{230,241,251}
\definecolor{bluedark}{RGB}{4,44,83}
\definecolor{blueaccent}{RGB}{181,212,244}

\definecolor{greenlight}{RGB}{234,243,222}
\definecolor{greendark}{RGB}{23,52,4}
\definecolor{greenaccent}{RGB}{192,221,151}

\definecolor{graylight}{RGB}{241,239,232}
\definecolor{graydark}{RGB}{44,44,42}
\definecolor{grayborder}{RGB}{136,135,128}

\definecolor{wrongred}{RGB}{153,60,29}
\definecolor{wrongbg}{RGB}{250,236,231}
\definecolor{rightgreen}{RGB}{59,109,17}
\definecolor{rightbg}{RGB}{234,243,222}

\title{IA-RAG: Interval-Algebra–Driven Temporal Reasoning for
Dynamic Knowledge Retrieval}

\author{
Xiaoman Wang\textsuperscript{1}\thanks{Equal contribution.},
Yaoze Zhang\textsuperscript{2,3}\footnotemark[1],
Wenzhuo Fan\textsuperscript{2,4},
Hongwei Zhang\textsuperscript{1,2},\\
\textbf{Ding Wang}\textsuperscript{2},
\textbf{Guohang Yan}\textsuperscript{2},
\textbf{Song Mao}\textsuperscript{2},
\textbf{Botian Shi}\textsuperscript{2},
\textbf{Yunshi Lan}\textsuperscript{1}\thanks{Corresponding author.},
\textbf{Pinlong Cai}\textsuperscript{2}\footnotemark[2]
\\
\textsuperscript{1}East China Normal University,
\textsuperscript{2}Shanghai Artificial Intelligence Laboratory, \\
\textsuperscript{3}University of Shanghai for Science and Technology, 
\textsuperscript{4}Harbin Engineering University \\
\texttt{xmwang@stu.ecnu.edu.cn}, \texttt{razzzhang@gmail.com}, \\
\texttt{yslan@dase.ecnu.edu.cn}, \texttt{caipinlong@pjlab.org.cn}
}

\begin{document}
\maketitle
\setlist{nosep}

% \begin{abstract}
% Retrieval-Augmented Generation (RAG) has shown strong effectiveness in grounding Large Language Models (LLMs) with external knowledge. However, existing RAG and Graph RAG frameworks largely treat knowledge as static or associate time with coarse-grained points or metadata, failing to capture the rich temporal topology of real-world events, including duration, overlap, and containment.  

% We propose \textbf{IA-RAG}, a hierarchical temporal RAG framework that explicitly models knowledge as time intervals and performs retrieval under formal temporal constraints. IA-RAG represents facts as \emph{Interval Event Units (IEUs)} and organizes them into a hierarchical \emph{Thematic Forest}, where temporal relationships are governed by Allen’s Interval Algebra. This structure enables explicit reasoning over complex temporal topologies beyond point-based time representations. To handle \textbf{incomplete or uncertain temporal boundaries}, IA-RAG introduces a \emph{Sub-graph Time Tightening} mechanism that refines fuzzy intervals by exploiting logical constraints within connected event subgraphs. Furthermore, IA-RAG supports \textbf{implicit temporal semantic retrieval} through a state-anchor-based strategy that resolves queries via interval-algebra–guided traversal.
% Experiments on temporal question answering benchmarks, including TimeQA, TempReason, and ComplexTR, demonstrate that IA-RAG consistently outperforms existing RAG and Graph RAG baselines in both retrieval precision and temporal reasoning accuracy.
% \end{abstract}

\begin{abstract}
% Retrieval-Augmented Generation (RAG) has shown strong effectiveness in grounding Large Language Models (LLMs) with external knowledge. However, existing RAG and Graph RAG frameworks largely treat knowledge as static or associate time with coarse-grained points or metadata, failing to capture the rich temporal topology of real-world events, including duration, overlap, and containment. We propose \textbf{IA-RAG}, a hierarchical temporal RAG framework that explicitly models knowledge as time intervals and performs retrieval under formal temporal constraints. IA-RAG represents facts as \emph{Interval Event Units (IEUs)} and organizes them into a hierarchical \emph{Thematic Forest}, where temporal relationships are governed by Allen’s Interval Algebra. This structure enables explicit reasoning over complex temporal topologies beyond point-based time representations. To handle \textbf{incomplete or uncertain temporal boundaries}, IA-RAG introduces a \emph{Sub-graph Time Tightening} mechanism that refines fuzzy intervals by exploiting logical constraints within connected event subgraphs. Furthermore, IA-RAG supports \textbf{implicit temporal semantic retrieval} through a state-anchor-based strategy that resolves queries via interval-algebra–guided traversal.
% \xmadd{Experiments on multiple temporal question answering benchmarks, including TimeQA, TempReason, and ComplexTR, demonstrate that IA-RAG achieves strong temporal retrieval and reasoning performance, particularly on complex compositional temporal reasoning tasks.}

Retrieval-Augmented Generation (RAG) has shown strong effectiveness in grounding Large Language Models (LLMs) with external knowledge. However, existing RAG and Graph RAG frameworks largely treat knowledge as static or associate time with coarse-grained timestamps or metadata, failing to capture rich temporal structures such as duration, overlap, and containment. We propose \textbf{IA-RAG}, a hierarchical temporal RAG framework that models knowledge as time intervals and performs retrieval under formal temporal constraints. IA-RAG represents facts as \emph{Interval Event Units (IEUs)} and organizes them into a hierarchical \emph{Thematic Forest}, where temporal dependencies are governed by Allen’s Interval Algebra. To handle incomplete or uncertain temporal boundaries, IA-RAG further introduces a \emph{Sub-graph Time Tightening} mechanism that refines fuzzy intervals through logical constraints within connected event subgraphs. In addition, IA-RAG supports implicit temporal semantic retrieval through interval-algebra-guided traversal. 
Experiments on multiple temporal question answering benchmarks, including TimeQA, TempReason, and ComplexTR, demonstrate that IA-RAG achieves strong temporal retrieval and reasoning performance, particularly on complex compositional temporal reasoning tasks.
Our code is released at \url{https://github.com/xiaoAugenstern/LogicalRAG_TemporalQA}.
% \yzcomment{Experimental results demonstrate that IA-RAG achieves state-of-the-art results on TimeQA and ComplexTR, and near state-of-the-art performance on TempReason, highlighting its effectiveness in complex compositional temporal reasoning.}

\end{abstract}

\section{Introduction}
\label{Introduction}
Large Language Models (LLMs) have achieved remarkable progress on knowledge-intensive tasks~\cite{naveed2025comprehensive}. However, they remain limited by outdated parametric knowledge and factual inconsistency. Retrieval-Augmented Generation (RAG) mitigates these issues by grounding generation in external evidence~\cite{lewis2021rag,zhang2025leanragknowledgegraphbasedgenerationsemantic}. More recently, Graph RAG methods extend this paradigm by modeling relational structures through entity-relation graphs~\cite{edge2025graphrag,guo2025lightrag}.

Despite these advances, existing RAG frameworks still lack an expressive mechanism for temporal reasoning. Recent temporal-aware approaches mainly incorporate timestamps through metadata filtering~\cite{li2025-t-grag}, temporal anchoring~\cite{sun2025dygragdynamicgraphretrievalaugmented}, or chronological organization~\cite{zhang2026respecting-e2rag}. However, real-world facts are not isolated time points; they are intervals with duration, overlap, and containment. Treating time as a shallow attribute leads to three major limitations:
(1)\textbf{Temporal Boundary Ambiguity}: Semantic retrievers struggle to distinguish between states of the same subject across different periods. For instance, queries regarding ``policies \textit{before} 2008" and ``policies \textit{after} 2008" yield nearly identical embeddings, causing the retrieval of temporally irrelevant facts. 
(2) \textbf{Inability to Refine Implicit Temporal Attributes}: Many real-world facts lack explicit timestamps. Current methods fail to utilize the complementary relationships between events to infer or impute these unknown temporal intervals, leaving a significant portion of the knowledge base temporally unanchored and unsearchable. 
(3) \textbf{Topological Reasoning Deficiency}: Complex queries often require inferring relationships between multiple events, such as determining if one event occurred during another or if they overlap. 
Existing RAG systems return disjoint chunks without modeling these interval-based dependencies, forcing LLMs to rely on brittle linguistic inference rather than structured temporal logic.

To bridge these gaps, we present IA-RAG, a hierarchical temporal RAG framework that leverages Allen’s Interval Algebra (IA) to rigorously model and reason over the topological structure of events. To systematically address the limitations of point-based approaches, our framework introduces three key innovations aligned with the identified challenges. First, to overcome topological reasoning deficiencies in existing event representations, we propose the \textbf{Interval Event Unit (IEU)} and the \textbf{Thematic Forest}. Unlike previous graphs that rely on shallow time-proximity links, our structure organizes atomic IEUs into a hierarchy where edges are governed by the 13 fundamental IA relations (e.g., \emph{meets}, \emph{overlaps}). Second, to tackle the inability to refine implicit temporal attributes, we introduce a \textbf{Sub-graph Time Tightening mechanism}. By exploiting the logical constraints within connected subgraphs, this mechanism infers missing timestamps and refines fuzzy boundaries based on the precise temporal context of neighboring events. Finally, to overcome temporal boundary ambiguity, IA-RAG employs an \textbf{Interval-Algebra–guided traversal strategy}. Instead of relying solely on semantic similarity, this approach performs directional retrieval anchored to specific event states, ensuring that queries are resolved with strict adherence to their temporal logic constraints.

Our contributions are summarized as follows:
\begin{itemize}[leftmargin=*]
    % 1.16上午更新   
    \item We introduce an interval-based temporal representation grounded in Allen’s Interval Algebra, which is the first to formally model event duration, topology, and temporal relations in retrieval-augmented generation, thereby enabling reasoning over intervals, in contrast to existing RAG methods that model time as discrete, unordered point timestamps.
    \item We propose IA-RAG, a hierarchical temporal GraphRAG framework that integrates logical temporal connectivity, ambiguity-aware boundary refinement, and constraint-guided retrieval. This enables accurate answering of implicit temporal state queries and interval-level reasoning tasks that are fundamentally inaccessible to conventional timestamp-based RAG systems.
    % \item Extensive experiments on challenging temporal QA benchmarks, including TimeQA, TempReason, and ComplexTR, demonstrate significant improvements in both retrieval precision and temporal reasoning accuracy over existing methods.
    \item Extensive experiments on multiple temporal QA benchmarks, including TimeQA, TempReason, and ComplexTR, show that IA-RAG consistently delivers strong temporal retrieval and reasoning performance, especially on complex compositional temporal reasoning tasks.
\end{itemize}

\begin{figure*}[h]
  \centering
  \includegraphics[width=0.88\textwidth]{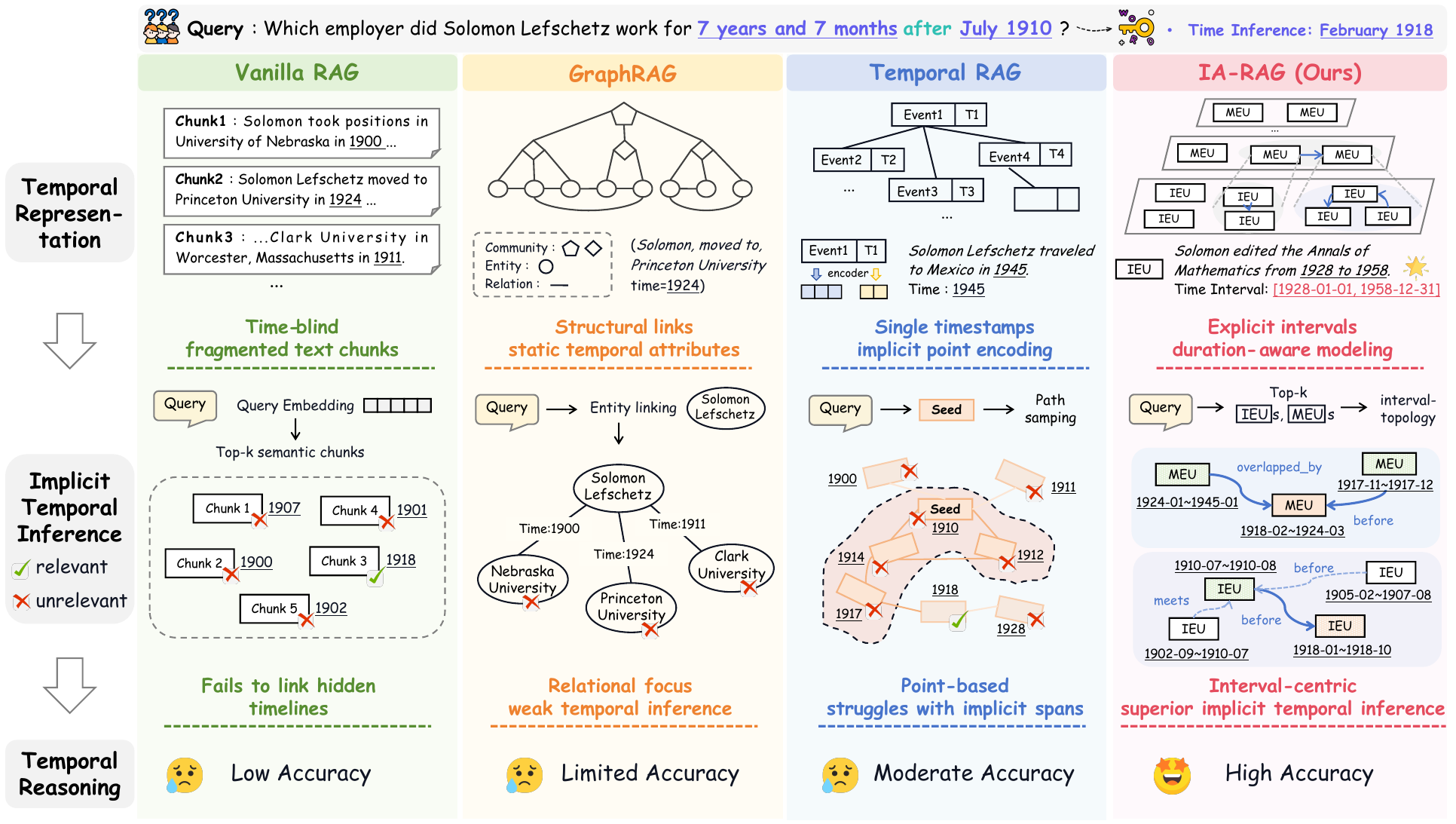}
  \caption{Comparison of representative RAG methods and IA-RAG.}
  \label{fig:framework}
\end{figure*}

\section{Related Work}
\label{related}

\subsection{Knowledge Graph Based Retrieval-Augmented Generation}

Retrieval-Augmented Generation (RAG) improves the factual reliability of Large Language Models (LLMs) by grounding generation in external evidence sources~\cite{lewis2021rag,yan2025hetaraghybriddeepretrievalaugmented}. To overcome the limitations of flat dense retrieval for multi-hop reasoning, graph-based RAG frameworks explicitly model relational structures among knowledge units. GraphRAG~\cite{edge2025graphrag} constructs entity-centric graphs for structured traversal, while HippoRAG~\cite{gutierrez2024hipporag} improves long-range relational recall through neuro-inspired memory mechanisms. Subsequent works such as LightRAG~\cite{guo2025lightrag} and MemoRAG~\cite{qian2025memorag} further enhance retrieval efficiency via hierarchical indexing and memory compression. 
Event-centric designs like EventRAG~\cite{yang-etal-2025-eventrag} extract events to capture logico-deductive narratives, and neuro-symbolic extensions like NeuSTIP~\cite{singh2023neustip} couple temporal graph completion with path-mining rules. 
Despite their advancements, these graph-based systems focus predominantly on static semantic relations. 
They relegate temporal dimensions to isolated node attributes or unlinked metadata, making them inherently incapable of capturing the topological variations of evolving facts. In contrast, IA-RAG introduces interval-aware event graphs and hierarchical thematic forests to support structured temporal retrieval.

\subsection{Temporal-Aware RAG}

Recent studies have explored incorporating temporal signals into the RAG pipeline for reasoning over evolving knowledge and time-sensitive facts~\cite{piryani2025itshightimesurvey}. TA-RAG~\cite{lau2025reading-ta-rag} addresses diachronic question answering by retrieving temporally coherent evidence across queried periods, while T-GRAG~\cite{li2025-t-grag} organizes historical knowledge into timestamped subgraphs to reduce temporal conflicts and redundancy. TG-RAG~\cite{han2025ragmeetstemporalgraphs} further introduces hierarchical temporal graphs for multi-granularity retrieval, and DyG-RAG~\cite{sun2025dygragdynamicgraphretrievalaugmented} models dynamic event-centric retrieval using temporally grounded event units. In parallel, $E^2$RAG~\cite{zhang2026respecting-e2rag} preserves evolving entity states through a dual entity-event graph architecture. However, existing temporal RAG systems primarily model time as discrete timestamps or temporal filters, lacking a formal semantic representation of interval relations such as overlap and duration. 
As a result, they struggle with queries involving implicit temporal states or complex interval constraints.
IA-RAG addresses this limitation by representing events as explicit temporal intervals and performing retrieval under Allen’s Interval Algebra~\cite{allen1984towards}, enabling structured reasoning over temporal topologies.
\section{Method}
\label{method}

\subsection{IA-RAG Workflow}
\label{sec:workflow}

\begin{figure*}[h]
  \centering
  \includegraphics[width=0.95\textwidth]{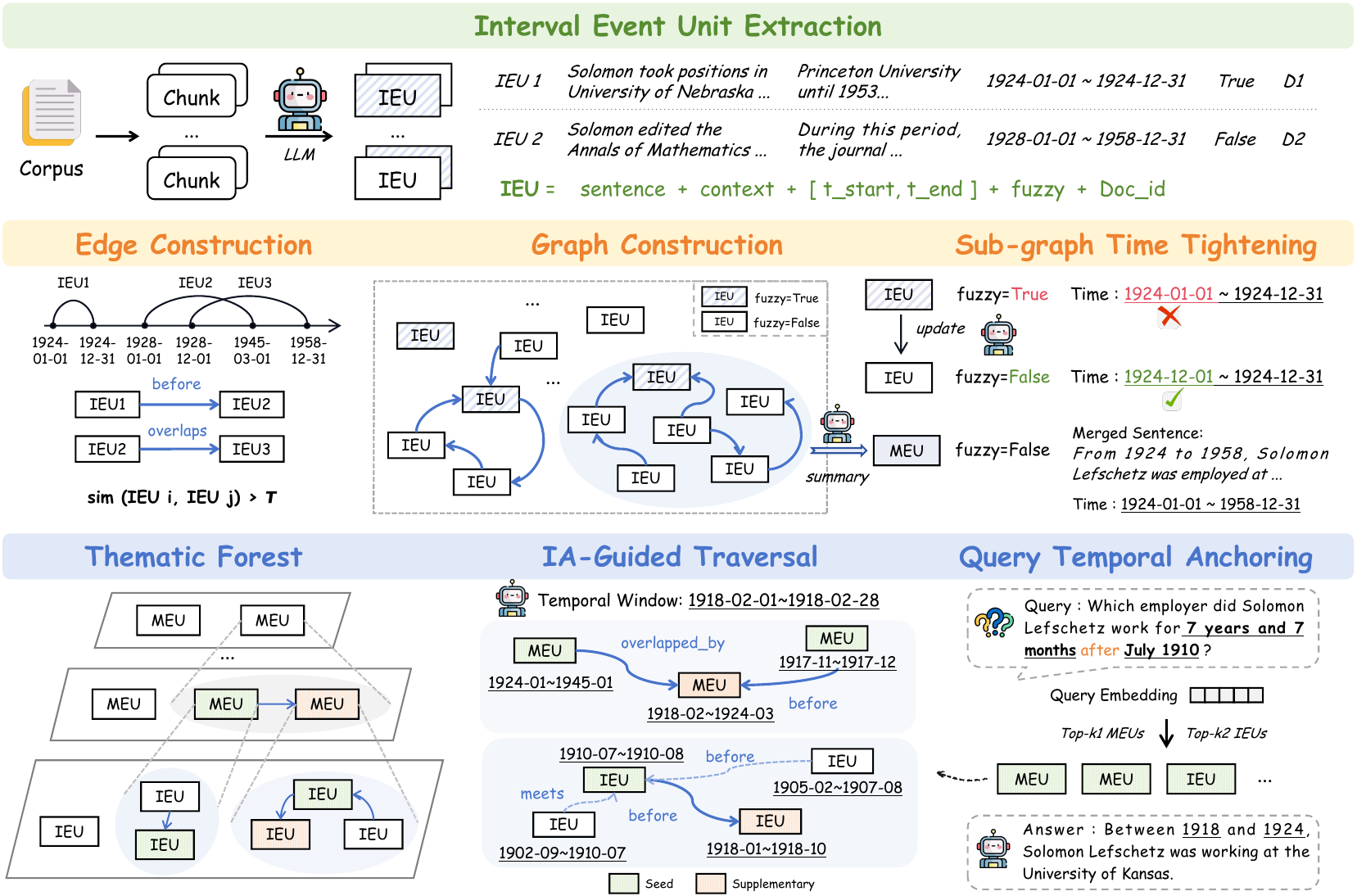}
  \caption{Overall framework of IA-RAG.}
  \label{fig:framework}
\end{figure*}

% In this section, we present \textbf{IA-RAG}, a temporal reasoning framework that extends Graph-based Retrieval-Augmented Generation by explicitly modeling events as intervals and performing retrieval under formal interval-algebraic constraints. The overall architecture of IA-RAG is illustrated in Figure~\ref{fig:framework}. The framework consists of three core stages: (1) interval event unit extraction, (2) interval-aware event graph construction, and (3) query-driven event retrieval.

In this section, we present \textbf{IA-RAG}, a temporal GraphRAG framework that models events as time intervals and performs retrieval under interval-algebraic constraints. As shown in Figure~\ref{fig:framework}, IA-RAG consists of three stages: (1) interval event unit extraction, (2) interval-aware event graph construction, and (3) query-driven temporal retrieval.

\subsection{Interval Event Unit Extraction}
\label{sec:ieu_extraction}

\paragraph{Definition of Interval Event Unit.} An IEU is a minimal and self-contained representation of a factual statement that describes either an instantaneous event or a persistent state over time. Formally, an IEU $e_i$ is defined as
\begin{equation}
e_i =
\langle
s_i,\,
c_i,\,
[t_i^{start},\,
t_i^{end}],\,
f_i,\,
src_i
\rangle,
\end{equation}
where $s_i$ denotes the event sentence, $c_i$ represents optional contextual information, $[t_i^{start}, t_i^{end}]$ specifies the temporal boundaries of the event, $f_i \in \{0,1\}$ is a temporal uncertainty indicator, and $src_{i}$ identifies the originating document.
% denoting whether the interval is inferred from imprecise or implicit temporal evidence

To extract IEUs, we first segment each source document into overlapping text chunks to preserve local coherence. 
Events with explicit timestamps are directly aligned to precise intervals with $f_i=0$, while vague expressions (e.g., ``early 1990s'') are mapped to coarse-grained intervals inferred from context and marked as uncertain with $f_i=1$.

% Candidate events are then identified within each chunk using syntactic patterns combined with temporal expression recognition. For events associated with explicit temporal references (e.g., a specific date or year), the temporal boundaries $t^{\text{start}}$ and $t^{\text{end}}$ are directly aligned with the corresponding timestamp, and the uncertainty flag is set to $f_i = 0$. In contrast, events described by vague or implicit temporal expressions (e.g., ``early 1990s'') are mapped to coarse-grained intervals inferred from contextual constraints, and the uncertainty indicator is activated $f_i = 1$. For instance, the expression ``early 1990s'' is normalized to the interval $[1990\text{-}01\text{-}01,\;1993\text{-}12\text{-}31]$, with $f_i = 1$. 

\paragraph{IEU Filtering and Deduplication}

The initial extraction phase inevitably yields redundant IEUs due to cross-document repetition and overlapping segmentation. To mitigate this redundancy, we perform IEU-level deduplication using a
\emph{local semantic neighborhood} strategy.
For each IEU $e_i$, we retrieve its top-$Q$ most semantically similar neighbors
$\mathcal{N}_Q(e_i)$ based on cosine similarity.
The set $\{e_i\} \cup \mathcal{N}_Q(e_i)$ is jointly examined by an LLM to determine whether a subset of IEUs represents duplicate descriptions of the same event. 
% If the evidence confirms that these IEUs are redundant descriptions of the same event, we consolidate them into a single, newly defined IEU assigned with a precise temporal interval $f=0$, thereby explicitly eliminating duplication. 
% \yzcomment{If the evidence confirms that these IEUs are redundant descriptions of the same event, we consolidate them into a single newly defined IEU and assign it a precise temporal interval inferred from the temporal spans of all constituent IEUs, thereby explicitly eliminating redundancy.}

\subsection{Interval-Aware Event Graph Construction}
\label{sec:graph_construction}
After filtering and deduplication, the extracted Interval Event Units (IEUs) serve as atomic nodes for temporal knowledge organization. We construct an \emph{interval-aware event graph} by explicitly modeling both semantic relatedness and qualitative temporal relations among IEUs.

\paragraph{Definition of Interval-Aware Edge.}
Given two IEUs $e_i$ and $e_j$, an interval-aware temporal relation is established
only when they satisfy both semantic relevance and temporal comparability.

Let $\mathbf{h}_i, \mathbf{h}_j \in \mathbb{R}^d$ denote dense semantic embeddings of the event descriptions. We first compute a semantic compatibility indicator:
\begin{equation}
\mathcal{S}(e_i, e_j)
=
\mathbb{I}
\left(
\frac{\mathbf{h}_i \cdot \mathbf{h}_j}
{\|\mathbf{h}_i\| \|\mathbf{h}_j\|}
>
\tau_{\text{sem}}
\right),
\end{equation}
where $\tau_{\text{sem}}$ is a predefined similarity threshold.
Only pairs with $\mathcal{S}(e_i, e_j)=1$ are considered for temporal relation assignment.

\paragraph{Allen Interval Relations.}
For semantically compatible IEUs, we determine their qualitative temporal relation using Allen’s Interval Algebra~\cite{allen1984towards}. Allen's Interval Algebra provides a rigorous formalism for modeling the relative ordering of time intervals, enabling the construction of a temporal topology among events. 
Given the temporal intervals
$[t_i^{start}, t_i^{end}]$ and
$[t_j^{start}, t_j^{end}]$,
their relation is defined as:
\begin{equation}
\mathcal{R}(e_i, e_j) \in \mathcal{A},
\end{equation}
where $\mathcal{A}$ denotes the set of 13 mutually exclusive Allen relations. Relations are detailed in Table~\ref{tab:allen_relations_full}.

\begin{table}[h]
\centering
\small
% 1. 使用 \resizebox 将表格宽度强制设置为 \columnwidth (单栏宽度)
% 2. 高度设置为 ! 表示自动保持长宽比
\resizebox{\columnwidth}{!}{%
    \begin{tabular}{l l l}
    \toprule
    \textbf{Relation} & \textbf{Definition} & \textbf{Inverse}\\
    \midrule
    \textit{before}         & $t_i^{end} < t_j^{start}$                         & \textit{after} \\
    \textit{meets}          & $t_i^{end} = t_j^{start}$                         & \textit{met\_by}\\
    \textit{overlaps}       & $t_i^{start} < t_j^{start} < t_i^{end} < t_j^{end}$ & \textit{overlapped\_by} \\
    \textit{starts}         & $t_i^{start} = t_j^{start}, t_i^{end} < t_j^{end}$ & \textit{started\_by}\\
    \textit{during}         & $t_j^{start} < t_i^{start} < t_i^{end} < t_j^{end}$ & \textit{contains}\\
    \textit{finishes}       & $t_i^{start} > t_j^{start}, t_i^{end} = t_j^{end}$ & \textit{finished\_by}\\
    \textit{equals}         & $t_i^{start} = t_j^{start}, t_i^{end} = t_j^{end}$ & -\\
    \bottomrule
    \end{tabular}%
} % resizebox 结束
\caption{The 13 mutually exclusive Allen's interval relations, including 7 basic relations and their corresponding inverses.}
\label{tab:allen_relations_full}
\end{table}

\paragraph{Sub-graph Time Tightening.}
% Contextual fragmentation caused by corpus-layer segmentation, together with the inherent sparsity of raw textual evidence, frequently leads to IEUs with highly uncertain temporal boundaries, which are marked by a fuzzy flag ($f=1$). Such temporal ambiguity degrades knowledge representation quality, causing affected IEUs to collapse into static and inefficient retrieval nodes, thereby severely limiting the effectiveness of the knowledge base. Leveraging the constructed temporal topology graph, IA-RAG mitigates this issue by recovering latent structural dependencies.

% To resolve temporal ambiguity, we first decompose the global event graph into structurally coherent substructures. 

Contextual fragmentation and sparse textual evidence often produce IEUs with uncertain temporal boundaries, marked by a fuzzy flag ($f=1$). To reduce such ambiguity, IA-RAG leverages the interval-aware event graph to recover latent temporal dependencies. Formally, we identify the set of connected components $\mathbf{C}$ via graph decomposition:
\begin{equation}
\mathbf{C} = \phi(G) = \left\{ \mathcal{C}_1, \mathcal{C}_2, \dots, \mathcal{C}_N \right\},
\end{equation}
where $\phi(\cdot)$ denotes the connected component computation. By incorporating Interval Algebra, discrete temporal signals are seamlessly bridged, allowing fragmented information to converge. This integration catalyzes the evolution of loose, theme-centric events into structured, interconnected components.

Based on this observation, given an IEU with the original temporal interval $\tau
\triangleq \left[ t^{start}, t^{end} \right]$ and the fuzzy status with $f=1$, we designate its encompassing component $\mathcal{C}_k \in \mathbf{C}$ as the grounded local context. Within $\mathcal{C}_k$, the surrounding IEUs and their explicit Allen relations impose logical constraints that implicitly bound the target's timeline. The refined temporal interval $\tau^*$ is then derived via an LLM-based reasoning operator $\mathcal{L}$:
\begin{equation}
\tau^* = \mathcal{L} \left( \tau \mid \mathcal{C}_k \right), 
\end{equation}Upon reasoning, if the contextual evidence in $\mathcal{C}_k$ supports a precise temporal alignment, the fuzzy flag is updated to $f=0$; otherwise, the IEU remains marked as fuzzy $f=1$.

\paragraph{Macro Event Units.} 
% Driven by the observation that connected components naturally cluster thematically aligned events, we capitalize on this emergent topological structure to introduce Macro Event Units (MEUs). By synthesizing the constituent events within each component, MEUs achieve a dual-purpose representation: they inherit the precise temporal expressiveness of atomic units while simultaneously facilitating high-layer thematic abstraction. Allen relations can be computed between temporal intervals at the lower layer, yielding an interval-aware graph at the higher layer of abstraction. This observation motivates a recursive hierarchical construction. \xmcomment{need to brief}
Connected components naturally group thematically related events. Based on this observation, we construct Macro Event Units (MEUs) by aggregating the events within each component.
MEUs preserve interval-aware temporal semantics while providing higher-level thematic abstraction. Allen relations are then computed between MEU intervals, producing a higher-level interval-aware graph and enabling recursive hierarchical construction.

We generalize this process by anchoring the base hierarchy to the initial global graph, defined as $\mathcal{G}^{(0)} \equiv G$ with component set $\widetilde{\mathcal{C}}^{(0)} \equiv \mathbf{C}$. Subsequently, this aggregation process for $l$ layer to $l+1$ layer is defined as:
\begin{equation}
\mathcal{U}^{(l+1)} = \left\{ g(\mathcal{C}_i) \mid \mathcal{C}_i \in \widetilde{\mathcal{C}}^{(l)} \right\},
\end{equation}
\begin{equation}
\mathcal{G}^{(l+1)} = \left(\mathcal{U}^{(l+1)}, \mathcal{E}^{(l+1)}\right),
\end{equation}
where 
%$\widetilde{\mathcal{C}}^{(0)}$ (equivalently denoted as $\mathbf{C}$ in the base layer) represents the set of connected components identified in the initial graph, and 
$g(\cdot)$ denotes a composite abstraction operator that unifies the temporal boundaries and summarizes the semantic content of the events within each component.
$\mathcal{G}^{(l+1)}$,  $\mathcal{U}^{(l+1)}$, and $\mathcal{E}^{(l+1)}$ represent the event graph, the component set of MEUs, and the component set of qualitative
temporal constraints among these MEUs at $l+1$ layer.
% This recursion terminates when the graph stabilizes or becomes disjoint.

\paragraph{Thematic Forest Construction.}
Combining the Eqs. (4)-(8), the recursive abstraction process induces a genealogical structure across layers. Specifically, each unit $u \in \mathcal{U}^{(l+1)}$ is synthesized from a unique component $e \in \mathcal{C}^{(l)}$, and a directed containment relationship exists $h(\cdot,\cdot)$ between the high-layer unit and its constituent events, defined as:
\begin{equation}
\mathcal{T}^{(l)} = h\left(u, e\right) \triangleq u \leftrightarrow e,
\end{equation}
where $e$ is an IEU when $l=0$ or a MEU when all $l>0$. Collectively, these relationships form a set of hierarchical trees, where the roots represent the most abstract narrative themes and the leaves correspond to atomic IEUs. We define this resulting multi-layered structure as the \textit{Thematic Forest}, which organizes temporal knowledge from granular details to broad narrative arcs.
% \yzadd{Such a hierarchical organization naturally supports multi-granularity retrieval and reasoning, enabling the system to flexibly access temporal knowledge at different levels of abstraction according to query requirements.}

\subsection{Query-Driven Event Retrieval}
\label{sec:temporal_retrieval}
\paragraph{Query Temporal Scope Anchoring.}
Given a user query $q$, our objective is to ground the query by inferring its temporal scope. To achieve this, we employ an LLM-based parser $\Phi_d(\cdot)$ to extract the corresponding temporal window:
\begin{equation}
T_w = [w_s, w_e] = \Phi_d(q),
\end{equation}

where $w_s$ and $w_e$ denote the inferred start and end timestamps.

\paragraph{Event-to-Forest Matching.}

Existing temporal RAG methods typically operate on flat graph structures, treating temporal information as isolated embedding attributes~\cite{sun2025dygragdynamicgraphretrievalaugmented}. Such designs overlook high-level thematic organization and often require expensive global traversal.

In contrast, IA-RAG adopts a hierarchical coarse-to-fine retrieval strategy over the proposed Thematic Forest, where IEUs provide fine-grained factual evidence, and MEUs capture high-level thematic and temporal context. This hierarchy enables more accurate retrieval while preserving both factual precision and temporal coherence.
Given a query $q$, we independently retrieve relevant MEUs and IEUs.:
\begin{equation}
\mathcal{M}^* =
\operatorname*{TopK_1}_{u \in \mathcal{U}}
\operatorname{Sim}(\mathbf{h}_q, \mathbf{h}_u),
\end{equation}
\begin{equation}
\mathcal{I}^* =
\operatorname*{TopK_2}_{e \in \mathcal{V}}
\operatorname{Sim}(\mathbf{h}_q, \mathbf{h}_e),
\end{equation}
where $\mathbf{h}_q$, $\mathbf{h}_u$, and $\mathbf{h}_e$ denote the embeddings of the query, MEUs, and IEUs. $\mathcal{U}$ and $\mathcal{V}$ denote the MEU and IEU spaces. $\mathcal{M}^*$ and $\mathcal{I}^*$ represent the retrieved thematic units and event units, respectively. $K_1$ and $K_2$ are retrieval hyperparameters\footnote{ Hyperparameter experiments are provided in the Appendix~\ref{app:sensitivity_analysis}}.

% The retrieved MEUs provide high-level thematic guidance, while the retrieved IEUs preserve fine-grained temporal evidence for subsequent interval-aware reasoning.

\paragraph{Interval-Algebra--Guided Traversal.}
Due to the absence of explicit temporal awareness, purely semantic retrieval may retrieve semantically relevant events that occurred at incorrect time periods.
To address this challenge, IA-RAG performs temporal traversal over both the MEU-level forest and the IEU-level event graph according to the relative position between retrieved intervals and the inferred query window $T_w$. For each retrieved unit $x_i \in \{\mathcal{M}^*, \mathcal{I}^*\}$ with temporal interval
\begin{equation}
T_i=[t_i^{start},t_i^{end}],
\end{equation}
the traversal direction is dynamically determined as:
\begin{equation}
\Psi(T_i,T_w)=
\begin{cases}
\mathcal{R}^{F}, & t_i^{end} < w_s,\\
\mathcal{R}^{B}, & t_i^{start} > w_e,\\
\mathcal{R}^{F}\cup\mathcal{R}^{B}, & \text{otherwise}.
\end{cases}
\end{equation}

Here, $\mathcal{R}^{F}$ and $\mathcal{R}^{B}$ denote the forward and backward Allen relation sets, respectively:
\begin{equation}
\begin{split}
\mathcal{R}^{F} = \{
\textit{before},
\textit{meets},
\textit{overlaps},\\
\textit{starts},
\textit{during},
\textit{finishes},
\textit{equals}
\},
\end{split}
\end{equation}
\begin{equation}
\begin{split}
\mathcal{R}^{B} =
\{
\textit{after},
\textit{met\_by},
\textit{overlapped\_by},\\
\textit{started\_by},
\textit{contains},
\textit{finished\_by},
\textit{equals}
\}.
\end{split}
\end{equation}

Traversal over both hierarchical MEU graphs and fine-grained IEU graphs proceeds forward, backward, or bidirectionally according to the Allen temporal relation between the retrieved unit and the target window: 
% \yzcomment{revised later}
\begin{equation}
\begin{split}
\mathcal{M}^{sup} = \Bigl\{
u_j \,\Bigm|\,
& \exists u_i \in \mathcal{M}^*,
\; u_j \in \mathcal{N}(u_i), \\
& \mathcal{R}(u_i,u_j)\in\Psi(T_i,T_w)
\Bigr\},
\end{split}
\end{equation}
\begin{equation}
\begin{split}
\mathcal{I}^{sup} = \Bigl\{
e_j \,\Bigm|\,
& \exists e_i \in \mathcal{I}^*,
\; e_j \in \mathcal{N}(e_i), \\
& \mathcal{R}(e_i,e_j)\in\Psi(T_i,T_w)
\Bigr\},
\end{split}
\end{equation}
where $\mathcal{N}(\cdot)$ denotes temporal neighbors connected through Allen interval relations. Finally, the supplementary units are merged with the initial retrieval results to form the final retrieval set:
\begin{equation}
\mathcal{I}^{target}
=
\mathcal{I}^*
\cup
\mathcal{I}^{sup}
\cup
\mathcal{M}^*
\cup
\mathcal{M}^{sup}.
\end{equation}

% \begin{equation}
% \mathcal{I}^{target}
% =
% \mathcal{I}^*
% \cup
% \mathcal{I}^{sup}
% \cup
% \mathcal{M}^{sup}.
% \end{equation}

% \yzadd{revision needed

\section{Experimental Setup}
\label{Experiments setup}
In our experiments, we aim to answer the following research questions:
% \begin{itemize}
%    \item  \textbf{RQ1:} How does IA-RAG compare against state-of-the-art baselines on \textbf{Temporal QA tasks}?(section xxx)
% \item \textbf{RQ2:} What is the contribution of the \textbf{Sub-graph Time Tightening} mechanism to resolving temporal ambiguity and improving retrieval precision?(section xx)

% \item \textbf{RQ3:} In what ways does the \textbf{hierarchical structural design} (Thematic Forest) facilitate the generation of high-quality answers?(Section xxx)
% \item \textbf{RQ4:} How effective is the \textbf{Interval-Algebra-guided traversal} in capturing complex temporal dependencies for directional reasoning? (Section xxx)
% % \item \textbf{RQ5:} Are \textbf{information-rich events} sufficient to independently sustain the question-answering process without external augmentation?   
% \end{itemize}

\begin{itemize}[leftmargin=*]
\item \textbf{RQ1:}
% How does IA-RAG compare with existing RAG and temporal retrieval baselines on temporal question answering benchmarks?
How does IA-RAG compare against state-of-the-art baselines on temporal QA tasks?
\item \textbf{RQ2:}
Can IA-RAG achieve superior temporal reasoning performance while reducing the amount of retrieved context?
\item \textbf{RQ3:}
% How do the hierarchical thematic forest and sub-graph time tightening mechanisms contribute to retrieval precision and answer quality?
How do the hierarchical forest and sub-graph time tightening modules affect retrieval and answer quality?
\item \textbf{RQ4:}
How effective is the proposed interval-algebra-guided traversal strategy for capturing directional temporal dependencies?
\item \textbf{RQ5:}
% How does IA-RAG perform across different categories of temporal reasoning, including static, relative, and arithmetic reasoning?
How does IA-RAG perform across different temporal reasoning types?
\end{itemize}

\subsection{Datasets and Evaluations}
% To evaluate the effectiveness of IA-RAG, we consider three representative categories of temporal questions: Implicit Temporal Inference, Event State Grounding, and Multi-hop Temporal Reasoning. 
We conduct experiments on three widely used temporal question answering benchmarks: \textbf{TimeQA}~\cite{chen2021dataset-timqa}, \textbf{TempReason}~\cite{tan2023bTempReason}, and \textbf{ComplexTR}~\cite{tan2024complextr}. 
% All three datasets are derived from large-scale Wikipedia-based corpora. 
Following the experimental setup of DyG-RAG~\cite{sun2025dygragdynamicgraphretrievalaugmented}, we directly adopt their processed document corpus and the corresponding question–answer pairs, ensuring a fair and controlled comparison. For evaluation, we follow the token-level evaluation protocol~\cite{zhou2025indepthanalysisgraphbasedrag}, reporting both Accuracy and Recall, computed based on the overlap between the predicted answers and the gold-standard answer.
\begin{table*}[!htbp]
\centering
% 1. 将默认的 6pt 列间距缩小为 3pt，让表格更紧凑
\setlength{\tabcolsep}{3pt}
% 2. 将表格整体字号缩小一级（\small 或 \footnotesize）
\small
\renewcommand{\arraystretch}{1.1}
\begin{tabularx}{\textwidth}{lXXXXXX}
\toprule
\multirow{2}{*}{\textbf{Methods}} 
& \multicolumn{2}{c}{\textbf{TimeQA}} 
& \multicolumn{2}{c}{\textbf{TempReason}} 
& \multicolumn{2}{c}{\textbf{ComplexTR}} \\
\cmidrule(lr){2-3}
\cmidrule(lr){4-5}
\cmidrule(lr){6-7}
& \textbf{Acc.} & \textbf{Recall} 
& \textbf{Acc.} & \textbf{Recall} 
& \textbf{Acc.} & \textbf{Recall} \\
\midrule
\multicolumn{7}{c}{\textit{Traditional RAG Methods}} \\
\midrule
Vanilla RAG     
& 56.33 & 64.98  
& 70.96 & 82.53  
& 55.01 & 68.68   \\
GraphRAG-L      
& 40.26 & 46.28 
& 56.11 & 67.55 
& 43.16 & 54.29 \\
GraphRAG-G      
& 10.10 & 13.83 
& 8.81 & 11.34 
& 20.97 & 30.81 \\
LightRAG-L      
& 34.94 & 39.89 
& 54.23 & 66.82 
& 41.03 & 49.46 \\
LightRAG-G      
& 6.66 & 8.12 
& 13.54 & 15.27 
& 12.46 & 13.56 \\
LightRAG-H      
& 36.36 & 43.06 
& 50.01 & 62.95 
& 42.68 & 53.59 \\
HippoRAG        
& 39.99 & 45.39 
& 69.80 & 80.54 
& 44.68 & 55.28 \\
E$^2$GraphRAG   
& 40.48 & 50.19 
& 61.29 & 73.58 
& 38.29 & 54.99 \\
\midrule
\multicolumn{7}{c}{\textit{Temporal QA Retrieval Methods}} \\
\midrule
T-GRAG
& 29.89 & 34.21
& 55.42 & 65.32
& 27.66 & 34.78 \\
TG-RAG
& 47.80 & 56.09
& 30.91 & 38.47
& 50.15 & 63.54 \\
TA-RAG          
& 42.25 & 48.92 
& 58.66 & 67.97
& 29.26 & 37.22 \\
DyG-RAG         
& \underline{58.78} & \underline{67.02} 
& \textbf{84.75} & \textbf{91.47} 
& \underline{55.62} & \underline{69.88} \\
\midrule
IA-RAG (ours)   
& \textbf{61.72} {\small\textcolor{teal}{(+2.94)}} 
& \textbf{69.25} {\small\textcolor{teal}{(+2.23)}}
& \underline{80.21}   
& \underline{89.64}
& \textbf{65.95} {\small\textcolor{teal}{(+10.33)}} 
& \textbf{77.42} {\small\textcolor{teal}{(+7.54)}} \\
\bottomrule
\end{tabularx}
\caption{
Overall performance comparison on temporal QA benchmarks. Acc. and Recall denote answer accuracy and retrieval recall, respectively. The best results are highlighted in \textbf{bold}, the second-best results are \underline{underlined}, and improvements over the strongest baseline are shown in \textcolor{teal}{teal}.
}
\label{tab:temporal_qa_results}
\end{table*}

\subsection{Baselines}
% To evaluate the effectiveness of IA-RAG, we compare our framework against a comprehensive suite of baselines, including vanilla RAG, Graph RAG methods, and the temporal-aware RAG framework.
We compare IA-RAG with representative Vanilla RAG, Graph RAG, and Temporal-Aware RAG methods:

\begin{itemize}[leftmargin=*]
% \item Vanilla RAG~\cite{lewis2021rag}: A standard retrieval baseline that utilises a vector database to store document chunks. It retrieves relevant text based on the semantic similarity between query embeddings and chunk embeddings.
\item Vanilla RAG~\cite{lewis2021rag}: Dense retrieval over vectorized document chunks.

% \item GraphRAG~\cite{edge2025graphrag}: This method structures retrieved information into communities and performs traversal to capture global context. We evaluate two variants: GraphRAG-L (local retrieval) and GraphRAG-G (global retrieval).
\item GraphRAG~\cite{edge2025graphrag}: Community-based graph retrieval with local (L) and global (G) variants.

% \item LightRAG~\cite{guo2025lightrag}: A framework that employs a dual-level indexing mechanism to capture both low-level entity details and high-level summaries. We include its local (L), global (G), and hybrid (H) retrieval variants.
\item LightRAG~\cite{guo2025lightrag}: Dual-level indexing framework with local (L), global (G), and hybrid (H) retrieval.

% \item HippoRAG~\cite{gutierrez2024hipporag}: This model mimics human long-term memory by constructing a knowledge graph and utilising personalised PageRank for multi-hop relational reasoning.
\item HippoRAG~\cite{gutierrez2024hipporag}: Mimics human memory by running personalized PageRank over knowledge graphs for multi-hop retrieval.

% \item E\textsuperscript{2}GraphRAG~\cite{zhao20252graphrag}: It constructs a summary tree and an entity graph based on document chunks and then constructs bidirectional indexes to capture their many-to-many relationships.
\item E\textsuperscript{2}GraphRAG~\cite{zhao20252graphrag}: Builds bidirectional indexes across a summary tree and an entity graph to capture many-to-many document relationships.

% \item TG-RAG~\cite{han2025ragmeetstemporalgraphs}: It constructs bi-level temporal graphs where facts are represented as timestamped edges, utilizing hierarchical time nodes to support multi-granularity summarization.
\item TG-RAG~\cite{han2025ragmeetstemporalgraphs}: Structures timestamped edges into bi-level graphs with hierarchical time nodes for multi-granularity summaries.

% \item T-GRAG~\cite{li2025-t-grag}: A dynamic GraphRAG framework that structures historical evolution into time-stamped subgraphs and utilizes an interactive, three-layer retriever to resolve knowledge conflicts and redundancy over time.
\item T-GRAG~\cite{li2025-t-grag}: Models historical evolution via timestamped subgraphs and a three-layer retriever to resolve knowledge conflicts.

% \item TA-RAG~\cite{lau2025reading-ta-rag}: A temporal framework tailored for diachronic queries that disentangles the question into a subject and a time window, employing a specialized retriever to calibrate semantic matching against continuous temporal coherence.
\item TA-RAG~\cite{lau2025reading-ta-rag}: Calibrates semantic matching against temporal coherence by separating queries into subjects and time windows.

\item DyG-RAG~\cite{sun2025dygragdynamicgraphretrievalaugmented}: It introduces dynamic event units and constructs a temporally grounded event graph to support time-aware retrieval and multi-hop temporal reasoning.

\end{itemize}

\subsection{Implementation Details}
We employ \texttt{Qwen2.5-14B-Instruct}~\cite{qwen2025qwen25technicalreport} as the large language model for all graph-construction and generation tasks. And all retrieval modules use the \texttt{BGE-M3}~\cite{chen2024bge} encoder. All documents are segmented using a sliding window approach with a fixed chunk size of 1,200 tokens and an overlap of 64 tokens. Hierarchical thematic forest is constructed with a maximum depth of 4 layers.  Additional implementation details, hyperparameter settings, and sensitivity analyses are provided in Appendix~\ref{app:implementation}.

% \begin{figure*}[h]  
%     \centering
%     \includegraphics[width=0.8\textwidth]{latex/img/temporal_qa_scatter.png} % \textwidth 语义更准确，代表整页文本宽度
%     \caption{Comparison between answer accuracy and prompt token consumption across temporal QA methods.}
%     \label{fig:placeholder}
% \end{figure*}

\section{Results and Analyses}
\subsection{Overall Performance Comparison (RQ1)}
\label{sec:overall_performance}
Table~\ref{tab:temporal_qa_results} reports the overall results on three temporal QA benchmarks.
Overall, IA-RAG achieves the best performance on TimeQA and ComplexTR, demonstrating the effectiveness of interval-algebra-guided traversal for complex temporal reasoning and multi-hop chronological inference. Compared with existing RAG and temporal retrieval baselines, IA-RAG consistently improves retrieval precision by explicitly modeling temporal interval relations and dynamically constraining traversal directions. 
On TempReason, DyG-RAG achieves slightly better results. We attribute this to the fact that many TempReason questions focus on single timestamp-oriented temporal matching, where DyG-RAG’s temporal-semantic embedding alignment and local graph expansion are particularly effective. In contrast, IA-RAG shows larger advantages on benchmarks requiring more complex directional temporal reasoning and interval-level dependency modeling.

\subsection{Efficiency Analysis (RQ2)}
\label{sec:efficiency_analysis}

\begin{figure}[!htbp]
    \centering
    \includegraphics[width=\columnwidth]{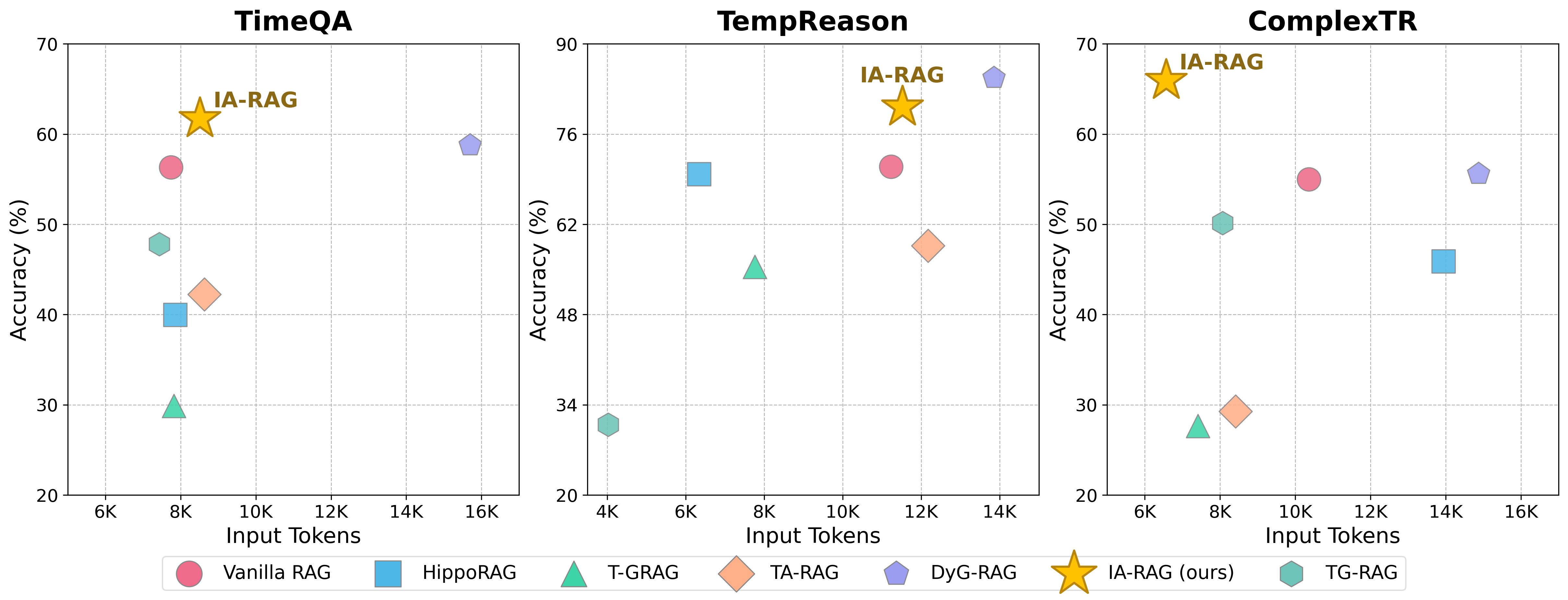}
    \caption{
    Comparison between answer accuracy and input token consumption across temporal QA methods.
    }
    \label{fig:temporal_qa_scatter}
\end{figure}

Figure~\ref{fig:temporal_qa_scatter} compares answer accuracy and input token consumption across different methods.
Although IA-RAG performs temporal traversal at both the MEU and IEU levels, it maintains relatively compact prompt sizes while achieving strong performance.
IA-RAG significantly reduces input tokens on ComplexTR while obtaining higher accuracy, indicating that the proposed interval-algebra-guided traversal retrieves more temporally relevant evidence instead of relying on exhaustive graph expansion.
Overall, the results demonstrate that IA-RAG achieves a better balance between retrieval effectiveness and efficiency for temporal QA.

\begin{table}[t]
\centering
\renewcommand{\arraystretch}{1.1}
\resizebox{\columnwidth}{!}{
\begin{tabular}{lcccccc}
\toprule
\multirow{2}{*}{\textbf{Methods}} 
& \multicolumn{2}{c}{\textbf{TimeQA}}
& \multicolumn{2}{c}{\textbf{TempReason}}
& \multicolumn{2}{c}{\textbf{ComplexTR}} \\
\cmidrule(lr){2-3} 
\cmidrule(lr){4-5} 
\cmidrule(lr){6-7}

& \textbf{Acc.} & \textbf{Recall} 
& \textbf{Acc.} & \textbf{Recall} 
& \textbf{Acc.} & \textbf{Recall}  \\
\midrule

\rowcolor{cyan!8}
IA-RAG       
& 61.72 & 69.25  
& 80.21 & 89.64
& 65.95 & 77.42 \\

\textit{w/o IA}           
& 57.90 & 65.48  
& 77.41 & 88.09
& 61.70 & 74.55 \\

\textit{w/o Tight.}    
& 60.81 & 68.66
& 79.66 & 89.12
& 63.52 & 76.34 \\

\textit{w/o Hier.}     
& 60.47 & 68.28 
& 78.98 & 88.44  
& 62.31 & 74.08 \\
\bottomrule
\end{tabular}
}
\caption{
Ablation study of IA-RAG. We evaluate the contribution of interval-algebra-guided traversal (\textit{IA}), hierarchical thematic retrieval (\textit{Hier.}), and sub-graph temporal tightening (\textit{Tight.}).
}
\label{tab:ablation}
\end{table}

% \subsection{Ablation Study (RQ3)}
\subsection{Effect of Core IA-RAG Components (RQ3)}
\label{sec:ablation_analysis}
Table~\ref{tab:ablation} presents the ablation results of IA-RAG across three benchmarks. 
Removing the interval-algebra-guided traversal (\textit{w/o IA}) precipitates the most severe performance drop, underscoring that pure semantic similarity is insufficient for complex temporal QA, which necessitates reasoning beyond isolated seed facts. 

Concurrently, removing the sub-graph time tightening mechanism (\textit{w/o Tight.}) consistently erodes accuracy; this confirms that factual evidence from fragmented corpora suffers from temporal boundary uncertainty, and tightening fuzzy intervals via local topological dependencies successfully eliminates temporal noise to preserve information-rich, chronologically coherent events. 
We further hypothesize that the slightly marginal gain from tightening on TempReason stems from the inherent temporal sensitivity bottlenecks of the backbone model in grounding explicit timestamps.

Removing the hierarchical structure (\textit{w/o Hier.}) harms performance, especially on ComplexTR, validating that our coarse-to-fine paradigm prevents retrieval from collapsing into flat event-level vector matching and reduces matching ambiguity.
% via global thematic organization.

\subsection{Traversal Strategy Analysis (RQ4)}
\label{sec:traversal_analysis}

Table~\ref{tab:traversal_ablation} compares different traversal strategies. Using all Allen relations without directional filtering leads to clear performance degradation, showing that unrestricted traversal introduces temporally irrelevant neighbors and noisy evidence. Replacing dynamic traversal with LLM-predicted top-$5$ Allen relations also underperforms IA-RAG.
Although LLM-predicted relations capture partial temporal dependencies, they often lack stable chronological consistency. In contrast, IA-RAG dynamically selects forward, backward, or bidirectional traversal according to the relative position between the seed event and the target temporal window.
This allows the model to focus on chronologically relevant regions while avoiding unnecessary expansion.

\begin{table}[t]
\centering
\renewcommand{\arraystretch}{1.1}

\resizebox{\columnwidth}{!}{
\begin{tabular}{lcccccc}
\toprule
\multirow{2}{*}{\textbf{Methods}} 
& \multicolumn{2}{c}{\textbf{TimeQA}}
& \multicolumn{2}{c}{\textbf{TempReason}}
& \multicolumn{2}{c}{\textbf{ComplexTR}} \\
\cmidrule(lr){2-3}
\cmidrule(lr){4-5}
\cmidrule(lr){6-7}
& \textbf{Acc.} & \textbf{Recall} 
& \textbf{Acc.} & \textbf{Recall} 
& \textbf{Acc.} & \textbf{Recall}  \\
\midrule

\rowcolor{cyan!8}
IA-RAG
& 61.72 & 69.25 
& 80.21 & 89.64 
& 65.95 & 77.42 \\

\textit{w/All}
& 60.54 & 68.89
& 77.49 & 88.10
& 62.91 & 75.75 \\

\textit{w/Top-5}
& 61.50 & 68.98
& 78.30 & 88.32
& 63.22 & 76.25 \\
\bottomrule
\end{tabular}
}
\caption{
Ablation study on traversal strategies.
\textit{All} denotes traversal over all Allen relations,
while \textit{Top-5} uses the top-5 Allen relations predicted by an LLM.
}
\label{tab:traversal_ablation}
\end{table}

\subsection{Breakdown Analysis on Temporal Reasoning Types (RQ5)}
\label{sec:breakdown_analysis}

%%%%%%%%%%%%%%%% break down %%%%%%%%%%%%%%%%%%%%
\begin{table}[t]
\centering
\renewcommand{\arraystretch}{1.15}
\setlength{\tabcolsep}{5pt}

\resizebox{\columnwidth}{!}{%
\begin{tabular}{lcccc}
\toprule
\textbf{Methods} 
& \textbf{Explicit}
& \textbf{Relative}
& \textbf{Comp.} 
& \textbf{Overall} \\
\midrule

TG-RAG
& 42.98
& \underline{68.49}
& 46.67
& 50.15 \\

Vanilla RAG      
& \underline{52.07}
& \underline{68.49} 
& 50.37  
& 55.01 \\

DyG-RAG          
& 51.24 
& 64.38 
& \underline{54.81}  
& \underline{55.62} \\
\midrule

\rowcolor{cyan!8}
\textbf{IA-RAG (Ours)} 
& \textbf{57.85} 
& \textbf{73.97} 
& \textbf{68.88} 
& \textbf{65.95} \\

\textit{Improvement}  
& {\textcolor{teal}{+5.78}} 
& {\textcolor{teal}{+5.48}} 
& {\textcolor{teal}{+14.07}} 
& {\textcolor{teal}{+10.33}} \\
\bottomrule
\end{tabular}%
}

\caption{
Accuracy breakdown on ComplexTR across different temporal reasoning categories.
}

\label{tab:complextr_breakdown}
\end{table}

Table~\ref{tab:complextr_breakdown} reports the performance breakdown on ComplexTR across different temporal reasoning categories\footnote{Detailed reasoning type definitions and dataset statistics are provided in Appendix~\ref{appendix:query_types}.}. IA-RAG consistently achieves the best performance across explicit, relative, and compositional temporal reasoning tasks.
The largest improvement is observed in compositional reasoning,
where IA-RAG surpasses the strongest baseline by more than 14\%, demonstrating the effectiveness of interval-algebra-guided traversal for multi-hop temporal dependency modeling and temporal computation. IA-RAG also shows clear gains in relative temporal reasoning,
indicating that dynamically constrained traversal effectively captures directional interval relations and event-conditioned temporal dependencies.

% \section{Conclusion}
% \label{conclusion}

% In this paper, we propose IA-RAG, a hierarchical temporal RAG framework that explicitly models events as time intervals and performs retrieval under Allen’s Interval Algebra constraints. By introducing Interval Event Units, interval-aware event graphs, sub-graph time tightening, and hierarchical thematic retrieval, IA-RAG enables structured reasoning over temporal overlap, containment, and ordering beyond conventional timestamp-based retrieval. Experimental results on multiple temporal QA benchmarks demonstrate that IA-RAG consistently improves both retrieval quality and temporal reasoning accuracy compared with existing RAG and GraphRAG baselines.

\section{Conclusion}
\label{conclusion}
% In this paper, we present IA-RAG, a hierarchical temporal RAG framework that models knowledge as time intervals and performs retrieval under Allen’s Interval Algebra constraints. By introducing Interval Event Units, interval-aware thematic forests, and sub-graph time tightening, IA-RAG enables structured temporal retrieval and reasoning beyond conventional timestamp-based approaches. In addition, the proposed interval-algebra-guided traversal supports implicit temporal semantic retrieval under complex temporal dependencies. Experiments on multiple temporal QA benchmarks demonstrate that IA-RAG achieves strong temporal retrieval and reasoning performance, particularly on complex compositional temporal reasoning tasks. 

In this paper, we present IA-RAG, a hierarchical temporal RAG framework that models knowledge as time intervals under Allen’s Interval Algebra. Through interval-aware retrieval, hierarchical thematic organization, and sub-graph time tightening, IA-RAG enables structured temporal reasoning beyond conventional timestamp-based retrieval. In addition, the proposed interval-algebra-guided traversal supports implicit temporal semantic retrieval under complex temporal dependencies. Experiments on multiple temporal QA benchmarks demonstrate strong performance, particularly on complex compositional temporal reasoning tasks.

% In this paper, we propose IA-RAG, a hierarchical temporal RAG framework that models events as time intervals and performs retrieval under Allen’s Interval Algebra constraints. By introducing interval-aware event graphs, temporal tightening, and hierarchical thematic retrieval, IA-RAG enables structured reasoning over temporal overlap, containment, and ordering beyond conventional timestamp-based retrieval. Experiments on multiple temporal QA benchmarks demonstrate consistent gains over existing RAG baselines.

% In this paper, we propose IA-RAG, a temporal RAG framework that models events as time intervals under Allen’s Interval Algebra. Through interval-aware retrieval and hierarchical reasoning, IA-RAG improves temporal reasoning beyond conventional timestamp-based retrieval. Experiments on multiple temporal QA benchmarks demonstrate consistent gains over existing RAG baselines.

\section*{Limitations}

Although IA-RAG achieves strong performance on temporal reasoning tasks, several limitations remain. First, the framework relies on the quality of temporal extraction and interval normalization, which may introduce errors when handling ambiguous or noisy temporal expressions. Second, the current framework mainly focuses on interval-level temporal reasoning and does not explicitly model causal relations or evolving event dynamics, which we leave for future work.

% \clearpage
\bibliography{main}

@inproceedings{tan2023bTempReason,
  title={Towards benchmarking and improving the temporal reasoning capability of large language models},
  author={Tan, Qingyu and Ng, Hwee Tou and Bing, Lidong},
  booktitle={Proceedings of the 61st Annual Meeting of the Association for Computational Linguistics (Volume 1: Long Papers)},
  pages={14820--14835},
  year={2023}
}

@inproceedings{tan2024complextr,
  title={Towards robust temporal reasoning of large language models via a multi-hop QA dataset and pseudo-instruction tuning},
  author={Tan, Qingyu and Ng, Hwee Tou and Bing, Lidong},
  booktitle={Findings of the Association for Computational Linguistics: ACL 2024},
  pages={6272--6286},
  year={2024}
}

@inproceedings{zhang2025leanragknowledgegraphbasedgenerationsemantic,
  title={Leanrag: Knowledge-graph-based generation with semantic aggregation and hierarchical retrieval},
  author={Zhang, Yaoze and Wu, Rong and Cai, Pinlong and Wang, Xiaoman and Yan, Guohang and Mao, Song and Wang, Ding and Shi, Botian},
  booktitle={Proceedings of the AAAI Conference on Artificial Intelligence},
  volume={40},
  number={41},
  pages={34862--34869},
  year={2026}
}

@article{sun2025dygragdynamicgraphretrievalaugmented,
  title={DyG-RAG: Dynamic Graph Retrieval-Augmented Generation with Event-Centric Reasoning},
  author={Sun, Qingyun and Yuan, Jiaqi and He, Shan and Guan, Xiao and Yuan, Haonan and Fu, Xingcheng and Li, Jianxin and Yu, Philip S},
  journal={arXiv preprint arXiv:2507.13396},
  year={2025}
}

@article{zhou2025indepthanalysisgraphbasedrag,
  title={In-depth Analysis of Graph-based RAG in a Unified Framework},
  author={Zhou, Yingli and Su, Yaodong and Sun, Youran and Wang, Shu and Wang, Taotao and He, Runyuan and Zhang, Yongwei and Liang, Sicong and Liu, Xilin and Ma, Yuchi and others},
  journal={arXiv preprint arXiv:2503.04338},
  year={2025}
}

@inproceedings{yang-etal-2025-eventrag,
  title={Eventrag: Enhancing llm generation with event knowledge graphs},
  author={Yang, Zairun and Wang, Yilin and Shi, Zhengyan and Yao, Yuan and Liang, Lei and Ding, Keyan and Yilmaz, Emine and Chen, Huajun and Zhang, Qiang},
  booktitle={Proceedings of the 63rd Annual Meeting of the Association for Computational Linguistics (Volume 1: Long Papers)},
  pages={16967--16979},
  year={2025}
}

@article{lewis2021rag,
  title={Retrieval-augmented generation for knowledge-intensive nlp tasks},
  author={Lewis, Patrick and Perez, Ethan and Piktus, Aleksandra and Petroni, Fabio and Karpukhin, Vladimir and Goyal, Naman and K{\"u}ttler, Heinrich and Lewis, Mike and Yih, Wen-tau and Rockt{\"a}schel, Tim and others},
  journal={Advances in neural information processing systems},
  volume={33},
  pages={9459--9474},
  year={2020}
}

@article{edge2025graphrag,
  title={From local to global: A graph rag approach to query-focused summarization},
  author={Edge, Darren and Trinh, Ha and Cheng, Newman and Bradley, Joshua and Chao, Alex and Mody, Apurva and Truitt, Steven and Metropolitansky, Dasha and Ness, Robert Osazuwa and Larson, Jonathan},
  journal={arXiv preprint arXiv:2404.16130},
  year={2024}
}

@article{guo2025lightrag,
  title={Lightrag: Simple and fast retrieval-augmented generation},
  author={Guo, Zirui and Xia, Lianghao and Yu, Yanhua and Ao, Tian and Huang, Chao},
  journal={arXiv preprint arXiv:2410.05779},
  volume={2},
  number={3},
  year={2024}
}

@article{zhao20252graphrag,
  title={E\^{} 2GraphRAG: Streamlining Graph-based RAG for High Efficiency and Effectiveness},
  author={Zhao, Yibo and Zhu, Jiapeng and Guo, Ye and He, Kangkang and Li, Xiang},
  journal={arXiv preprint arXiv:2505.24226},
  year={2025}
}

@article{gutierrez2024hipporag,
  title={Hipporag: Neurobiologically inspired long-term memory for large language models},
  author={Guti{\'e}rrez, Bernal J and Shu, Yiheng and Gu, Yu and Yasunaga, Michihiro and Su, Yu},
  journal={Advances in neural information processing systems},
  volume={37},
  pages={59532--59569},
  year={2024}
}

@inproceedings{qian2025memorag,
  title={Memorag: Boosting long context processing with global memory-enhanced retrieval augmentation},
  author={Qian, Hongjin and Liu, Zheng and Zhang, Peitian and Mao, Kelong and Lian, Defu and Dou, Zhicheng and Huang, Tiejun},
  booktitle={Proceedings of the ACM on Web Conference 2025},
  pages={2366--2377},
  year={2025}
}

@article{han2025ragmeetstemporalgraphs,
  title={RAG Meets Temporal Graphs: Time-Sensitive Modeling and Retrieval for Evolving Knowledge},
  author={Han, Jiale and Cheung, Austin and Wei, Yubai and Yu, Zheng and Wang, Xusheng and Zhu, Bing and Yang, Yi},
  journal={arXiv preprint arXiv:2510.13590},
  year={2025}
}

@article{chen2021dataset-timqa,
  title={A dataset for answering time-sensitive questions},
  author={Chen, Wenhu and Wang, Xinyi and Wang, William Yang},
  journal={arXiv preprint arXiv:2108.06314},
  year={2021}
}

@inproceedings{li2025-t-grag,
  title={T-grag: A dynamic graphrag framework for resolving temporal conflicts and redundancy in knowledge retrieval},
  author={Li, Dong and Niu, Yichen and Ai, Ying and Zou, Xiang and Qi, Biqing and Liu, Jianxing},
  booktitle={Proceedings of the 33rd ACM International Conference on Multimedia},
  pages={11880--11889},
  year={2025}
}

@article{lau2025reading-ta-rag,
  title={Reading Between the Timelines: RAG for Answering Diachronic Questions},
  author={Lau, Kwun Hang and Zhang, Ruiyuan and Shi, Weijie and Zhou, Xiaofang and Cheng, Xiaojun},
  journal={arXiv preprint arXiv:2507.22917},
  year={2025}
}

@misc{qwen2025qwen25technicalreport,
      title={Qwen2.5 Technical Report}, 
      author={Qwen and An Yang and Baosong Yang and Beichen Zhang and Binyuan Hui and Bo Zheng and Bowen Yu and Chengyuan Li and Dayiheng Liu and Fei Huang and Haoran Wei and Huan Lin and Jian Yang and Jianhong Tu and Jianwei Zhang and Jianxin Yang and Jiaxi Yang and Jingren Zhou and Junyang Lin and Kai Dang and Keming Lu and Keqin Bao and Kexin Yang and Le Yu and Mei Li and Mingfeng Xue and Pei Zhang and Qin Zhu and Rui Men and Runji Lin and Tianhao Li and Tianyi Tang and Tingyu Xia and Xingzhang Ren and Xuancheng Ren and Yang Fan and Yang Su and Yichang Zhang and Yu Wan and Yuqiong Liu and Zeyu Cui and Zhenru Zhang and Zihan Qiu},
      year={2025},
      eprint={2412.15115},
      archivePrefix={arXiv},
      primaryClass={cs.CL},
     url={https://arxiv.org/abs/2412.15115}, 
}

@article{chen2024bge,
  title={{BGE M3}-embedding: Multi-lingual, multi-functionality, multi-granularity text embeddings through self-knowledge distillation},
  author={Chen, Jianlv and Xiao, Shitao and Zhang, Peitian and Luo, Kun and Lian, Defu and Liu, Zheng},
  journal={arXiv preprint arXiv:2402.03216},
  volume={4},
  number={5},
  year={2024}
}

@article{allen1984towards,
  title={Towards a general theory of action and time},
  author={Allen, James F},
  journal={Artificial intelligence},
  volume={23},
  number={2},
  pages={123--154},
  year={1984}
}

@article{naveed2025comprehensive,
  title={A comprehensive overview of large language models},
  author={Naveed, Humza and Khan, Asad Ullah and Qiu, Shi and Saqib, Muhammad and Anwar, Saeed and Usman, Muhammad and Akhtar, Naveed and Barnes, Nick and Mian, Ajmal},
  journal={ACM Transactions on Intelligent Systems and Technology},
  volume={16},
  number={5},
  pages={1--72},
  year={2025}
}

@misc{piryani2025itshightimesurvey,
      title={It's High Time: A Survey of Temporal Question Answering}, 
      author={Bhawna Piryani and Abdelrahman Abdallah and Jamshid Mozafari and Avishek Anand and Adam Jatowt},
      year={2025},
      eprint={2505.20243},
      archivePrefix={arXiv},
      primaryClass={cs.CL},
      url={https://arxiv.org/abs/2505.20243}, 
}

@misc{yan2025hetaraghybriddeepretrievalaugmented,
      title={HetaRAG: Hybrid Deep Retrieval-Augmented Generation across Heterogeneous Data Stores}, 
      author={Guohang Yan and Yue Zhang and Pinlong Cai and Ding Wang and Song Mao and Hongwei Zhang and Yaoze Zhang and Hairong Zhang and Xinyu Cai and Botian Shi},
      year={2025},
      eprint={2509.21336},
      archivePrefix={arXiv},
      primaryClass={cs.IR},
      url={https://arxiv.org/abs/2509.21336}, 
}

@article{singh2023neustip,
  title={Neustip: A novel neuro-symbolic model for link and time prediction in temporal knowledge graphs},
  author={Singh, Ishaan and Kaur, Navdeep and Gaur, Garima and others},
  journal={arXiv preprint arXiv:2305.11301},
  year={2023}
}

@inproceedings{zhang2026respecting-e2rag,
  title={Respecting Temporal-Causal Consistency: Entity-Event Knowledge Graph for Retrieval-Augmented Generation},
  author={Zhang, Ze Yu and Li, Zitao and Li, Yaliang and Ding, Bolin and Low, Bryan Kian Hsiang},
  booktitle={Proceedings of the 19th Conference of the European Chapter of the Association for Computational Linguistics (Volume 1: Long Papers)},
  pages={2017--2054},
  year={2026}
}

\appendix

\clearpage

\section{Use of Large Language Models}
The research presented in this paper, including the core ideas, experimental design, and quantitative results, is the original work of the authors. A large language model was used as a writing assistant for tasks such as polishing prose, improving clarity, and correcting grammatical errors in the manuscript. All final content was reviewed and edited by the authors to ensure it accurately reflects our research and contributions.

\section{Experimental Setup}
\label{app:implementation}

\subsection{Setting}

\paragraph{Model and Inference.}
All stages of our pipeline, including Interval Event Unit (IEU) extraction, semantic filtering, deduplication, and question answering, are implemented using \texttt{Qwen2.5-14B-Instruct}\footnote{\url{https://huggingface.co/Qwen/Qwen2.5-14B-Instruct}}.

\paragraph{Hardware Configuration.}
All experiments were conducted on a machine equipped with
4 $\times$ NVIDIA A100 GPUs (80GB memory per GPU).
Inference is performed in parallel where applicable.

\paragraph{Similarity Thresholds.}
For IEU-level semantic filtering and deduplication, we compute cosine similarity between event embeddings and apply a similarity threshold of $\tau_{\text{sem}} = 0.8$. Only event pairs exceeding this threshold are considered semantically related.

\paragraph{Thematic Forest Construction.}
During Thematic Forest construction, we apply a semantic similarity threshold
of $0.8$ at the first abstraction layer ($\mathcal{U}^{(0)} \rightarrow \mathcal{U}^{(1)}$)
to form Macro Event Units (MEUs).
The same threshold is used consistently across subsequent layers unless otherwise specified.

\paragraph{Retrieval Configuration.}

In the retrieval stage, we adopt a two-step coarse-to-fine retrieval strategy.
Specifically, we first retrieve the top-$K_1$ most relevant MEUs, followed by the top-$K_2$ IEUs. For TimeQA and ComplexTR, we set $K_1=10$, while $K_1=20$ is used for TempReason. Across all datasets, we fix $K_2=20$.

\begin{table}[h]
\centering
\small
\begin{tabularx}{\columnwidth}{l l X}
\toprule
\textbf{Parameter} & \textbf{Value} & \textbf{Description} \\
\midrule
$K$ & 50 & Number of nearest neighbors \\
$T_{\text{sem}}$ & 0.8 & Semantic similarity threshold \\
MIN\_CLUSTER\_SIZE & 2 & Minimum size for aggregation \\
MAX\_LEVELS & 4 & Maximum hierarchy depth \\
DECAY & 0.95 & Threshold decay factor \\
MIN\_THRESHOLD & 0.6 & Lower bound of similarity threshold \\
MAX\_THRESHOLD & 0.85 & Initial similarity threshold \\
MAX\_WORKERS & 32 & Parallel processing threads \\
\bottomrule
\end{tabularx}
\caption{Key hyperparameters used in graph construction.}
\label{tab:hyperparams}
\end{table}

\subsection{Sensitivity Analysis on Forest Depth and MEU Granularity}
\label{app:sensitivity_analysis}

\begin{table*}[t]
\centering

\renewcommand{\arraystretch}{1.1}

\begin{tabularx}{\textwidth}{lXXXXXX}
\toprule

\multirow{2}{*}{\textbf{MAX\_LEVELS}} 
& \multicolumn{2}{c}{\textbf{TimeQA}} 
& \multicolumn{2}{c}{\textbf{TempReason}} 
& \multicolumn{2}{c}{\textbf{ComplexTR}} \\

\cmidrule(lr){2-3}
\cmidrule(lr){4-5}
\cmidrule(lr){6-7}

& \textbf{Acc.} & \textbf{Recall} 
& \textbf{Acc.} & \textbf{Recall} 
& \textbf{Acc.} & \textbf{Recall} \\

\midrule

0 
&60.47 & 68.28
&78.98 & 88.44
&62.31 & 74.08 \\

4 
& \textbf{61.72} & \textbf{69.25}
& \textbf{80.21} & \textbf{89.64}
& \textbf{65.95} & \textbf{77.42} \\

6 
& 60.70 & 68.49 
& 77.67 & 87.89
& 62.31 & 75.70\\

10 
& 60.19 & 68.18
& 77.49 & 87.88
& 63.52 & 76.54\\

\bottomrule
\end{tabularx}
\caption{
Sensitivity analysis of thematic forest depth (\texttt{MAX\_LEVELS}).
\texttt{MAX\_LEVELS}=0 corresponds to a flat graph without hierarchical thematic organization.
}
\label{tab:max_level}
\end{table*}

\begin{table*}[t]
\centering

\renewcommand{\arraystretch}{1.1}
\begin{tabularx}{\textwidth}{lXXXXXX}
\toprule

\multirow{2}{*}{\textbf{MEU}} 
& \multicolumn{2}{c}{\textbf{TimeQA}} 
& \multicolumn{2}{c}{\textbf{TempReason}} 
& \multicolumn{2}{c}{\textbf{ComplexTR}} \\

\cmidrule(lr){2-3}
\cmidrule(lr){4-5}
\cmidrule(lr){6-7}

& \textbf{Acc.} & \textbf{Recall} 
& \textbf{Acc.} & \textbf{Recall} 
& \textbf{Acc.} & \textbf{Recall} \\

\midrule
5
& 59.85 & 67.81
& 77.10 & 87.53
& 63.22 & 74.64 \\

10 
&\textbf{61.72} & \textbf{69.25}
& 76.52 & 87.48
&\textbf{65.95} & \textbf{77.42}\\

15 
& 60.93 & 68.47
& 77.91 & 88.19
& 62.91 & 75.89 \\

20 
& 59.89 & 67.64
& \textbf{80.21} & \textbf{89.64}
& 60.79 & 74.04  \\

40 
& 59.51 & 67.55
& 76.52 & 86.23
& 60.79 & 74.49\\

\bottomrule
\end{tabularx}
\caption{
Sensitivity analysis of MEU retrieval granularity.
MEU denotes the number of retrieved thematic units used for coarse-to-fine retrieval.
}
\label{tab:meu_topk}
\end{table*}

Tables~\ref{tab:max_level} and~\ref{tab:meu_topk} present the sensitivity analysis of IA-RAG with respect to thematic forest depth (\texttt{MAX\_LEVELS}) and MEU retrieval granularity.

For thematic forest depth, setting \texttt{MAX\_LEVELS}=0 degenerates IA-RAG into a flat graph without hierarchical thematic abstraction, resulting in consistent performance drops across all benchmarks. Introducing moderate hierarchical depth significantly improves both accuracy and recall, indicating that multi-level thematic organization effectively narrows the retrieval space before fine-grained event reasoning. However, excessively deep hierarchies slightly degrade performance, likely because over-fragmented thematic partitions weaken semantic coherence and increase retrieval sparsity.

For MEU retrieval granularity, a moderate number of retrieved MEUs achieves the best overall performance. Retrieving too few MEUs limits thematic coverage and may overlook relevant temporal evidence, while retrieving too many introduces noisy or weakly related subgraphs that reduce retrieval precision. These results demonstrate that IA-RAG benefits from a balanced coarse-to-fine retrieval strategy that preserves thematic diversity while maintaining temporal relevance.

% \begin{figure*}[h]  
%     \centering
%     \includegraphics[width=0.9\textwidth]{latex/img/temporal_qa_scatter.png} % \textwidth 语义更准确，代表整页文本宽度
%     \caption{Comparison between answer accuracy and prompt token consumption across temporal QA methods.}
%     \label{fig:placeholder}
% \end{figure*}

\section{Complexity and Theoretical Analysis}
\label{app:appendix_complexity}
% Let $N$ denote the number of atomic events and $K$ the number of nearest neighbors per event. At level 0, each event retrieves its top-$K$ neighbors using an ANN index, resulting in a practical complexity of $O(N\log N)$. Since each node connects to at most $K$ neighbors, the number of edges is $O(NK)$, and temporal relation computation for each edge is constant time. Therefore, the overall complexity of level-0 graph construction is:
% \begin{equation}
% O(N\log N + NK) \approx O(N\log N)
% \end{equation}

% For hierarchical construction, each level performs connected component detection, component summarization, and edge reconstruction. Assuming the number of nodes decreases geometrically across levels, i.e.,
% \begin{equation}
% N_{l+1} \leq \alpha N_l, \quad 0 < \alpha < 1
% \end{equation}
% the cumulative graph construction cost across all levels remains approximately:
% \begin{equation}
% O(N\log N)
% \end{equation}

% In addition, the hierarchical summarization stage requires LLM calls for each connected component. Let $C_l$ denote the number of components at level $l$, and $T_{\text{LLM}}$ the average cost of one summarization call. The total complexity of the framework is therefore:
% \begin{equation}
% O\left(N\log N + \sum_l C_l \cdot T_{\text{LLM}}\right)
% \end{equation}

Let $N$ denote the number of atomic events and $K$ the number of nearest neighbors retrieved for each event. 
The overall complexity of the framework can be decomposed into three stages: retrieval, graph construction, and hierarchical summarization.

\paragraph{Retrieval Complexity.}
At level 0, each atomic event retrieves its top-$K$ nearest neighbors using an approximate nearest neighbor (ANN) index. 
Assuming ANN retrieval requires approximately $O(\log N)$ time per query, the total retrieval complexity is:
\begin{equation}
T_{\text{retrieval}} = O(N \log N)
\end{equation}

\paragraph{Graph Construction Complexity.}
After retrieval, each node connects to at most $K$ neighbors, resulting in at most $O(NK)$ edges. 
Temporal relation computation for each edge is constant time, and connected component detection can be performed in linear time with respect to the number of nodes and edges. 
Therefore, the graph construction complexity at each level is:
\begin{equation}
T_{\text{graph}} = O(NK)
\end{equation}

For hierarchical graph construction, let $N_l$ denote the number of nodes at level $l$. 
Assuming the graph size decreases geometrically across levels:
\begin{equation}
N_{l+1} \leq \alpha N_l, \quad 0 < \alpha < 1
\end{equation}
the cumulative graph construction cost across all levels becomes:
\begin{equation}
\sum_l O(N_l K) = O(NK)
\end{equation}

\paragraph{Hierarchical Summarization Complexity.}
At each hierarchy level, connected components are summarized using an LLM. 
Let $C_l$ denote the number of connected components at level $l$, and let $t_{\text{LLM}}$ denote the average inference cost of one summarization call. 
The total summarization complexity is therefore:
\begin{equation}
T_{\text{LLM}} = \sum_l C_l \cdot t_{\text{LLM}}
\end{equation}

\paragraph{Overall Complexity.}
Combining all stages, the overall complexity of the framework is:
\begin{equation}
T_{\text{total}}
=
O(N \log N)
+
O(NK)
+
\sum_l C_l \cdot t_{\text{LLM}}
\end{equation}

Since $K$ is a small constant in practice, the dominant non-LLM cost is the ANN-based retrieval stage with approximate complexity:
\begin{equation}
T_{\text{total}}
\approx
O(N \log N)
+
\sum_l C_l \cdot t_{\text{LLM}}
\end{equation}

\section{Temporal Query Type Definition and Dataset Distribution}
% \xmcomment{type name corresponding with section5.5 , revise later }
\label{appendix:query_types}

To systematically analyze temporal reasoning complexity, we categorize temporal questions using an LLM-based taxonomy classifier (\texttt{Qwen2.5-14B-Instruct}). We organize the queries into three high-level reasoning types, which encompass five fine-grained categories:

\begin{description}[leftmargin=!, font=\normalfont\bfseries]
    \item[Explicit Temporal Reasoning]
    These queries contain explicit timestamps or intervals, primarily requiring direct temporal matching. This group includes:
    \begin{itemize}[nosep, leftmargin=*, label=$\bullet$]
        \item \textbf{T1: Time-Anchored.} Queries with explicit timestamps or fixed ranges. \\
        \textit{Ex: ``Which position did Brian Smith hold after May 2001?''}
        \item \textbf{T2: Explicit Interval.} Queries with clearly specified start and end times. \\
        \textit{Ex: ``Which team did Attaphol Buspakom play for between May 1989 and Oct 1990?''}
    \end{itemize}

    % --- Group 2:  Reasoning ---
    \item[Relative Temporal Reasoning]
    These queries define time implicitly through event relations, requiring external temporal grounding. This group includes:
    \begin{itemize}[nosep, leftmargin=*, label=$\bullet$]
        \item \textbf{T4: Event-to-Event Relative.} Queries defining temporal relations relative to another event. \\
        \textit{Ex: ``Where was Beate Meinl-Reisinger educated before being a member of NEOS?''}
    \end{itemize}

    % --- Group 3:  Computation ---
    \item[Compositional Temporal Reasoning]
    These queries involve temporal arithmetic or multi-hop reasoning over both events and durations. This group includes:
    \begin{itemize}[nosep, leftmargin=*, label=$\bullet$]
        \item \textbf{T3: Numerical Offset.} Queries involving arithmetic operations over explicit timestamps. \\
        \textit{Ex: ``Which employer did Gustav Ludwig Hertz work for 8 years before April 1926?''}
        \item \textbf{T5: Multi-hop Reasoning.} Queries combining event relations with duration-based computation. \\
        \textit{Ex: ``Which employer did Manuel García Velarde work for 3 years after working for Université libre de Bruxelles?''}
    \end{itemize}
\end{description}

As shown in Table~\ref{tab:query_type_distribution}, ComplexTR contains substantially more relative and composite temporal reasoning questions than TimeQA and TempReason, making it significantly more challenging for retrieval-based temporal QA systems.

\begin{table}[t]
\centering
\small
\renewcommand{\arraystretch}{1.1}

\begin{tabular*}{\columnwidth}{@{\extracolsep{\fill}}lcccccc@{}}
\toprule
\textbf{Dataset}
& \textbf{T1}
& \textbf{T2}
& \textbf{T3}
& \textbf{T4}
& \textbf{T5}
& \textbf{Total} \\
\midrule

TimeQA
& 1589
& 1007
& 10
& 7
& 0
& 2613 \\

TempReason
& 5397
& 0
& 0
& 0
& 0
& 5397 \\

ComplexTR
& 72
& 49
& 42
& 73
& 93
& 329 \\

\bottomrule
\end{tabular*}

\caption{
Distribution of fine-grained temporal reasoning types across datasets.
}

\label{tab:t1-t5}
\end{table}

\begin{table}[t]
\centering
\small
\renewcommand{\arraystretch}{1.1}

\begin{tabular*}{\columnwidth}{@{\extracolsep{\fill}}lccc@{}}
\toprule
\textbf{Dataset}
& \textbf{Explicit}
& \textbf{Relative}
& \textbf{Compositional} \\
\midrule

TimeQA
& 2596
& 7
& 10 \\

TempReason
& 5397
& 0
& 0 \\

ComplexTR
& 121
& 73
& 135 \\

\bottomrule
\end{tabular*}

\caption{
Distribution of high-level temporal reasoning types across datasets.
% Explicit reasoning includes fixed timestamp and interval queries (T1--T2),
% Relative reasoning involves event-conditioned temporal grounding (T4),
% and Compositional reasoning requires temporal arithmetic or multi-hop inference (T3--T5).
}

\label{tab:query_type_distribution}
\end{table}

\section{Distribution of Allen Interval Relations Across Temporal QA Benchmarks}
\label{appendix:allen_distribution}
\begin{figure*}[h]
  \centering
  \includegraphics[width=\textwidth]{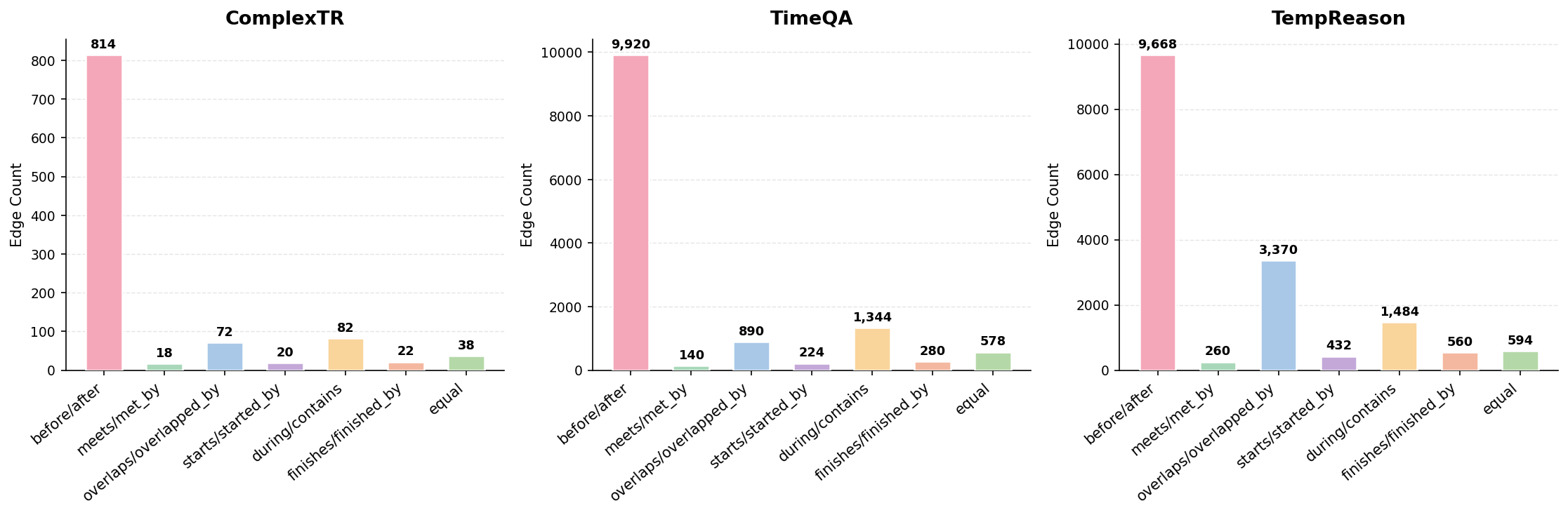}
  \caption{Distribution of Allen interval relations across temporal QA benchmarks. Inverse relation pairs are merged for clarity (e.g., \textit{before}/\textit{after}, \textit{during}/\textit{contains}).}
  \label{fig:allen_distribution}
\end{figure*}

To better understand the temporal topology characteristics of existing temporal QA benchmarks, we analyze the distribution of Allen interval relations within the constructed interval-aware event graphs. Figure~\ref{fig:allen_distribution} presents the relation distributions across TimeQA, TempReason, and ComplexTR.

% Specifically, for each pair of semantically related Interval Event Units (IEUs), 
% we compute their qualitative temporal relation according to Allen’s Interval Algebra. Following the standard formulation, we group inverse relations together 
% (e.g., \textit{before}/\textit{after}, \textit{during}/\textit{contains}) to highlight the overall structural composition of temporal dependencies.

First, although chronological relations such as 
\textit{before}/\textit{after} dominate all datasets, 
a substantial number of non-trivial interval relations also exist, including \textit{overlaps}, \textit{during}/\textit{contains}, 
and \textit{meets}. This indicates that temporal reasoning in real-world QA benchmarks extends beyond simple timestamp ordering. Second, the existence of diverse interval topologies empirically justifies the necessity of interval-aware retrieval and traversal mechanisms. These findings support our motivation that modeling time as isolated timestamps is insufficient for capturing the rich temporal dependencies present in complex QA tasks.

\section{Visualization of Hierarchical Thematic Forest}

\label{sec:forest_visualization}
\begin{figure*}[h]
  \centering
  \includegraphics[width=0.95\textwidth]{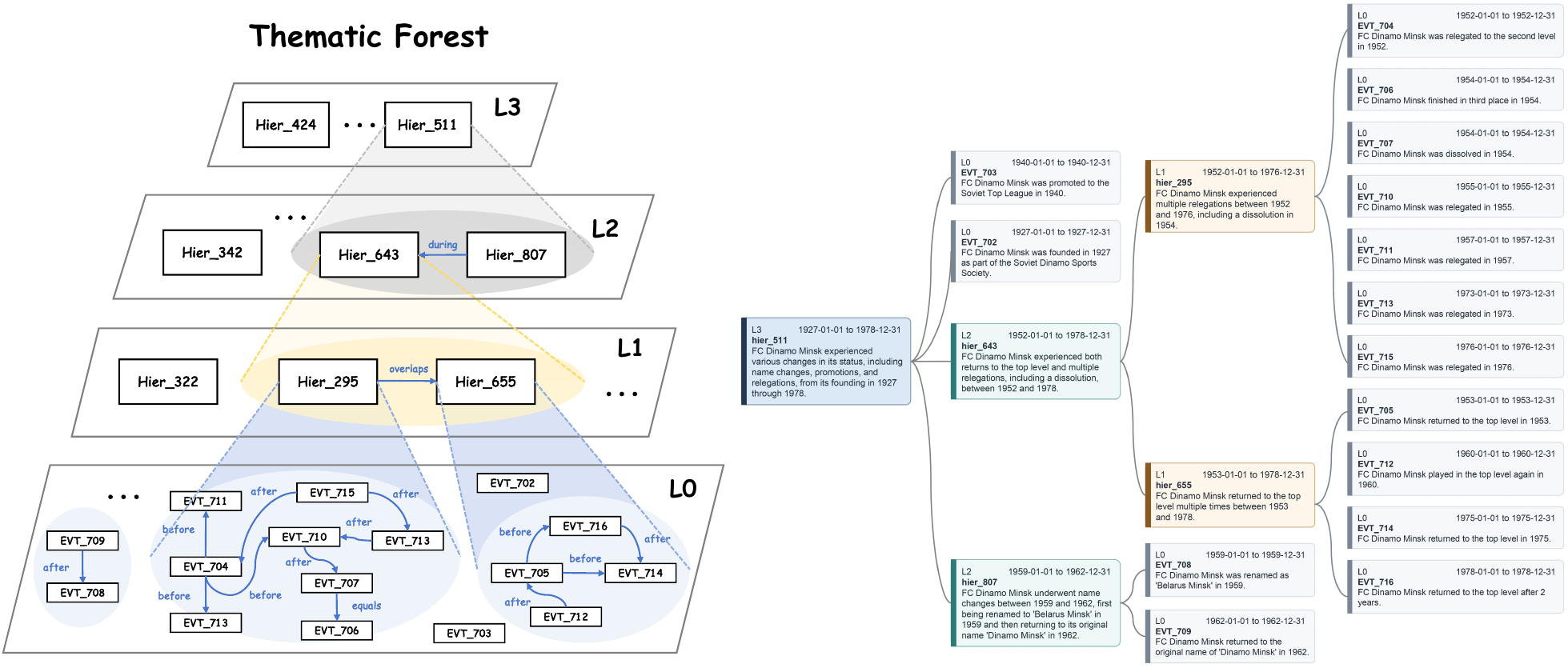}
  \caption{Visualization of the hierarchical thematic forest constructed by IA-RAG on ComplexTR.}
  \label{fig:forest_cropped}
\end{figure*}

To further illustrate the hierarchical organization mechanism of IA-RAG, we visualize representative thematic forest structures constructed on the ComplexTR benchmark. Figure~\ref{fig:forest_cropped} presents the hierarchical relationship between Macro Event Units (MEUs) and atomic Interval Event Units (IEUs).

At the lower level, IEUs represent fine-grained temporally grounded events connected through Allen interval relations. IA-RAG further aggregates semantically and temporally coherent IEUs into higher-level MEUs, which capture broader narrative themes and temporal subtopics. Through recursive abstraction, the resulting thematic forest organizes temporal knowledge in a coarse-to-fine manner, enabling efficient retrieval and directional temporal reasoning.

The visualization demonstrates that IA-RAG can simultaneously preserve detailed event-level temporal structure while constructing high-level thematic abstractions, thereby improving retrieval localization and reducing temporal ambiguity during multi-hop reasoning.

\section{Clarification on TG-RAG Performance on TempReason}

During experimental reproduction, we observed that a portion of the events in the TempReason dataset contain missing temporal annotations represented as \texttt{UNKNOWN}. This issue directly affects the graph construction mechanism of TG-RAG, which relies on explicit temporal information to establish temporal connections between events. To ensure a fair comparison, we excluded nodes with \texttt{UNKNOWN} timestamps during graph construction and conducted experiments on the resulting graph, from which the reported performance was obtained. We further conjecture that the relatively weak performance of TG-RAG on TempReason may stem from the limited robustness of its graph construction strategy under incomplete temporal supervision, particularly when temporal information is missing or ambiguously specified.

\section{Case Study}

To further illustrate the effectiveness of IA-RAG, we present several representative case studies in Tables~\ref{tab:case_study_complextr_T2}, \ref{tab:case_study_complextr_T4}, and \ref{tab:case_study_complextr_T3}. These examples demonstrate how IA-RAG performs temporal interval grounding under complex temporal reasoning scenarios, while existing RAG baselines often retrieve temporally irrelevant or logically inconsistent evidence.

% --------------------- Case study 2 -------------

\begin{table*}[t]
\centering
\small

% ── 问题框 ──────────────────────────────────────────────────────────────────
\begin{mdframed}[
  backgroundcolor=graylight,
  linecolor=grayborder,
  linewidth=1.5pt,
  topline=false, rightline=false, bottomline=false,
  leftline=true,
  innerleftmargin=10pt, innerrightmargin=10pt,
  innertopmargin=8pt, innerbottommargin=8pt,
  skipabove=0pt, skipbelow=6pt,
]
{\fontsize{9}{11}\selectfont\color{grayborder}\textbf{QUESTION:}}
{\small\bfseries\color{graydark}
Which employer did Karl Menger work for 3 years and 2 months before February 1930?}\\
{\fontsize{9}{11}\selectfont\color{grayborder} \textbf{Gold answer:}\quad}
{\fontsize{9}{11}\selectfont\bfseries\color{graydark} University of Amsterdam}
\end{mdframed}

% ── (a) Chunk-based ─────────────────────────────────────────────────────────
\begin{mdframed}[
  backgroundcolor=amberlight,
  linewidth=0pt,
  innerleftmargin=10pt, innerrightmargin=10pt,
  innertopmargin=8pt, innerbottommargin=8pt,
  skipabove=0pt, skipbelow=6pt,
  roundcorner=4pt,
]
{\small
\textbf{\color{amberdark}(a) Chunk-based Answer:}\hfill
\colorbox{wrongbg}{\fontsize{8}{10}\selectfont\color{wrongred}\textbf{$\times$ Incorrect}}

\medskip
{\color{amberdark}
Based on the information provided in the text chunks, Karl Menger worked at the University of Vienna from 1927 to 1930. Given the timeline provided, Karl Menger worked at the University of Vienna for approximately 2 years and 3 months before February 1930. Therefore, the employer he worked for 3 years and 2 months before February 1930 does not align exactly with any single employment stint mentioned. However, the closest match is his position at the University of Vienna, which lasted less than the specified duration. The question might contain an error or assumption that doesn't fit perfectly with the given facts about Karl Menger's career timeline.
}}
\end{mdframed}

% ── (b) DyG-RAG ─────────────────────────────────────────────────────────────
\begin{mdframed}[
  backgroundcolor=bluelight,
  linewidth=0pt,
  innerleftmargin=10pt, innerrightmargin=10pt,
  innertopmargin=8pt, innerbottommargin=8pt,
  skipabove=0pt, skipbelow=6pt,
  roundcorner=4pt,
]
{\small
\textbf{\color{bluedark}(b) DyG-RAG Answer:}\hfill
\colorbox{wrongbg}{\fontsize{8}{10}\selectfont\color{wrongred}\textbf{$\times$ Incorrect}}

\medskip
{\color{bluedark}
To determine Karl Menger's employer 3 years and 2 months before February 1930, we need to look back to December 1926. Based on the provided events and text chunks, we can analyze the timeline:

\smallskip
\textbf{Events Analysis}
\begin{itemize}[leftmargin=1.2em, itemsep=1pt, topsep=2pt]
  \item Event \#10 states that Karl Menger accepted a professorship at the University of Vienna in 1927.
  \item Event \#11 mentions Hertz becoming ordinarius professor of experimental physics and director of the Physics Institute of the Technische Hochschule Berlin in 1928, which is irrelevant to Menger's employment.
  \item Event \#12 talks about Albert Einstein nominating Heisenberg, Born, and Jordan for the Nobel Prize in Physics in 1928, which does not relate to Menger's employment.
  \item Event \#20 notes that Karl Menger was a visiting lecturer at Harvard University and The Rice Institute in 1930 and 1931. Given this information, we can deduce that Menger was not at Harvard University or The Rice Institute prior to 1930.
\end{itemize}

\textbf{Conclusion}

From the timeline and the events provided, Karl Menger was at the University of Vienna from 1927 to 1929. Thus, 3 years and 2 months before February 1930 would place us in December 1926, during which time Menger was still at the University of Vienna.

\smallskip
\textbf{Answer:} The employer of Karl Menger 3 years and 2 months before February 1930 was the University of Vienna.
}}
\end{mdframed}

% ── (c) IA-RAG ──────────────────────────────────────────────────────────────
\begin{mdframed}[
  backgroundcolor=greenlight,
  linewidth=0pt,
  innerleftmargin=10pt, innerrightmargin=10pt,
  innertopmargin=8pt, innerbottommargin=8pt,
  skipabove=0pt, skipbelow=6pt,
  roundcorner=4pt,
]
{\small
\textbf{\color{greendark}(c) IA-RAG Answer:}\hfill
\colorbox{rightbg}{\fontsize{8}{10}\selectfont\color{rightgreen}\textbf{$\checkmark$ Correct}}

\medskip
{\color{greendark}
To determine Karl Menger's employer 3 years and 2 months before February 1930, we need to identify his employment status in October 1926.

\smallskip
\textbf{Timeline Analysis}
\begin{itemize}[leftmargin=1.2em, itemsep=1pt, topsep=2pt]
  \item Event \#1 (1924): Karl Menger received his PhD from the University of Vienna.
  \item \begin{mdframed}[
  backgroundcolor=greenaccent,
  linewidth=0pt,
  innerleftmargin=0pt, innerrightmargin=8pt,
  innertopmargin=5pt, innerbottommargin=5pt,
  skipabove=2pt, skipbelow=2pt,
  roundcorner=3pt,
]
 Event \#2 (1925): L.~E.~J.\ Brouwer invited Karl Menger to teach at the University of Amsterdam in 1925.
\end{mdframed}
 
  \item Event \#3 (1927): Karl Menger returned to Vienna to accept a professorship there in 1927.
\end{itemize}

\textbf{Conclusion}

According to the timeline, Menger taught at the University of Amsterdam from 1925 to 1927, before returning to Vienna. Therefore, in October 1926, Karl Menger was employed at the University of Amsterdam. This inference is consistent with the information in Chunk 1.

\smallskip
\textbf{Answer:} Karl Menger worked for the University of Amsterdam 3 years and 2 months before February 1930.
}}
\end{mdframed}

\caption{Case study 1}
\label{tab:case_study_complextr_T2}
\end{table*}

\begin{table*}[t]
\centering
\small

% ── 问题框 ──────────────────────────────────────────────────────────────────
\begin{mdframed}[
  backgroundcolor=graylight,
  linecolor=grayborder,
  linewidth=1.5pt,
  topline=false, rightline=false, bottomline=false,
  leftline=true,
  innerleftmargin=10pt, innerrightmargin=10pt,
  innertopmargin=8pt, innerbottommargin=8pt,
  skipabove=0pt, skipbelow=6pt,
]
{\fontsize{9}{11}\selectfont\color{grayborder}\textbf{QUESTION}}\\[2pt]
{\small\bfseries\color{graydark}
Where was Stefan Mazurkiewicz educated after he/she studied at Ludwig Maximilian University of Munich?}\\[4pt]
{\fontsize{9}{11}\selectfont\color{grayborder} Gold answer:\quad}
{\fontsize{9}{11}\selectfont\bfseries\color{graydark} University of Göttingen}
\end{mdframed}

% ── (a) Chunk-based ─────────────────────────────────────────────────────────
\begin{mdframed}[
  backgroundcolor=amberlight,
  linewidth=0pt,
  innerleftmargin=10pt, innerrightmargin=10pt,
  innertopmargin=8pt, innerbottommargin=8pt,
  skipabove=0pt, skipbelow=6pt,
  roundcorner=4pt,
]
{\small
\textbf{\color{amberdark}(a) Chunk-based Answer:}\hfill
\colorbox{wrongbg}{\fontsize{8}{10}\selectfont\color{wrongred}\textbf{$\times$ Incorrect}}

\medskip
{\color{amberdark}
Based on the information provided in the text chunks, Stefan Mazurkiewicz, after studying at Ludwig Maximilian University of Munich, became a professor at the University of Paris for a time. However, the text does not specify the exact educational institution he attended immediately after Ludwig Maximilian University of Munich. It is clear that he spent most of his career as a professor at the University of Warsaw. Therefore, the immediate educational step after Ludwig Maximilian University of Munich mentioned is his period as a professor at the University of Paris, though it is not specified if he pursued further formal education there or elsewhere before taking up that position.
}}
\end{mdframed}

% ── (b) DyG-RAG ─────────────────────────────────────────────────────────────
\begin{mdframed}[
  backgroundcolor=bluelight,
  linewidth=0pt,
  innerleftmargin=10pt, innerrightmargin=10pt,
  innertopmargin=8pt, innerbottommargin=8pt,
  skipabove=0pt, skipbelow=6pt,
  roundcorner=4pt,
]
{\small
\textbf{\color{bluedark}(b) DyG-RAG Answer:}\hfill
\colorbox{wrongbg}{\fontsize{8}{10}\selectfont\color{wrongred}\textbf{$\times$ Incorrect}}

\medskip
{\color{bluedark}
None of the provided events or text chunks contain information about Stefan Mazurkiewicz's educational background after Ludwig Maximilian University of Munich. Therefore, the question cannot be answered based on the given data. The information provided focuses on individuals and events unrelated to Stefan Mazurkiewicz's educational history.
}}
\end{mdframed}

% ── (c) IA-RAG ──────────────────────────────────────────────────────────────
\begin{mdframed}[
  backgroundcolor=greenlight,
  linewidth=0pt,
  innerleftmargin=10pt, innerrightmargin=10pt,
  innertopmargin=8pt, innerbottommargin=8pt,
  skipabove=0pt, skipbelow=6pt,
  roundcorner=4pt,
]
{\small
\textbf{\color{greendark}(c) IA-RAG Answer:}\hfill
\colorbox{rightbg}{\fontsize{8}{10}\selectfont\color{rightgreen}\textbf{$\checkmark$ Correct}}

\medskip
{\color{greendark}
Based on the information provided in the text chunks and events, Stefan Mazurkiewicz was educated at the University of Göttingen after studying at Ludwig Maximilian University of Munich.

\smallskip
From the text chunk:

\begin{mdframed}[
  backgroundcolor=greenaccent,
  linewidth=0pt,
  innerleftmargin=8pt, innerrightmargin=8pt,
  innertopmargin=5pt, innerbottommargin=5pt,
  skipabove=2pt, skipbelow=2pt,
  roundcorner=3pt,
]
\textit{``Stefan Mazurkiewicz was a Polish mathematician who worked in mathematical analysis, topology, and probability. He was a student of Wacław Sierpiński and a member of the Polish Academy of Learning (``PAU'').''}
\end{mdframed}

\smallskip
And specifically from Event 3:

\begin{mdframed}[
  backgroundcolor=greenaccent,
  linewidth=0pt,
  innerleftmargin=8pt, innerrightmargin=8pt,
  innertopmargin=5pt, innerbottommargin=5pt,
  skipabove=2pt, skipbelow=2pt,
  roundcorner=3pt,
]
\textit{``He then went to the University of Göttingen on a grant from the Rockefeller Foundation to study mathematics under David Hilbert.''}
\end{mdframed}

\smallskip
Thus, after studying at Ludwig Maximilian University of Munich, Mazurkiewicz proceeded to the University of Göttingen to further his education in mathematics.
}}
\end{mdframed}

\caption{Case study 2}
\label{tab:case_study_complextr_T4}
\end{table*}

\begin{table*}[t]
\centering
\small

% ── 问题框 ──────────────────────────────────────────────────────────────────
\begin{mdframed}[
  backgroundcolor=graylight,
  linecolor=grayborder,
  linewidth=1.5pt,
  topline=false, rightline=false, bottomline=false,
  leftline=true,
  innerleftmargin=10pt, innerrightmargin=10pt,
  innertopmargin=8pt, innerbottommargin=8pt,
  skipabove=0pt, skipbelow=6pt,
]
{\fontsize{9}{11}\selectfont\color{grayborder}\textbf{QUESTION:}}
{\small\bfseries\color{graydark}
Where was Mary, Crown Princess of Denmark educated 15 years and 6 months before he/she was married to Frederik, Crown Prince of Denmark?}\\[4pt]
{\fontsize{9}{11}\selectfont\color{grayborder} Gold answer:\quad}
{\fontsize{9}{11}\selectfont\bfseries\color{graydark} Hobart College}
\end{mdframed}

% ── (a) Chunk-based ─────────────────────────────────────────────────────────
\begin{mdframed}[
  backgroundcolor=amberlight,
  linewidth=0pt,
  innerleftmargin=10pt, innerrightmargin=10pt,
  innertopmargin=3pt, innerbottommargin=3pt,
  skipabove=0pt, skipbelow=6pt,
  roundcorner=4pt,
]
{\small
\textbf{\color{amberdark}(a) Chunk-based Answer:}\hfill
\colorbox{wrongbg}{\fontsize{8}{10}\selectfont\color{wrongred}\textbf{$\times$ Incorrect}}

\medskip
{\color{amberdark}
Based on the provided text chunks, Mary, Crown Princess of Denmark was educated at the University of Tasmania from 1990 to 1994, where she graduated with a combined Bachelor of Commerce and Bachelor of Laws degree on 27 May 1995.\\
To determine where she was educated 15 years and 6 months before her marriage to Frederik, Crown Prince of Denmark, we need to calculate the dates:\\
Mary and Frederik got engaged on 8 October 2003 and were married on 14 May 2004.\\
15 years and 6 months before her marriage (14 May 2004) would be 14 November 1988.\\
Given this timeframe, we can see that during 14 November 1988, Mary was in her final year of high school at Taroona High School in Hobart, Tasmania, as she was completing her secondary schooling from 1984 to 1987 and matriculation from 18 to 1989. Therefore, 15 years and 6 months before her marriage, Mary was attending Taroona High School in Hobart, Tasmania.
}}
\end{mdframed}

% ── (b) DyG-RAG ─────────────────────────────────────────────────────────────
\begin{mdframed}[
  backgroundcolor=bluelight,
  linewidth=0pt,
  innerleftmargin=10pt, innerrightmargin=10pt,
  innertopmargin=3pt, innerbottommargin=3pt,
  skipabove=0pt, skipbelow=6pt,
  roundcorner=4pt,
]
{\small
\textbf{\color{bluedark}(b) DyG-RAG Answer:}\hfill
\colorbox{wrongbg}{\fontsize{8}{10}\selectfont\color{wrongred}\textbf{$\times$ Incorrect}}

\medskip
{\color{bluedark}
To determine where Mary, Crown Princess of Denmark, was educated 15 years and 6 months before her marriage to Frederik, Crown Prince of Denmark, we need to calculate the relevant timeline:\\
\textbf{Timeline Calculation}\\
Event \#16 indicates that Mary and Frederik were married on 14 May 2004.\\
15 years and 6 months before 14 May 2004 would be 14 May 1988. \\
\textbf{Education Analysis}\\
From the provided text chunks, specifically Chunk 7:
\begin{itemize}[leftmargin=1.2em, itemsep=1pt, topsep=2pt]
  \item Mary Elizabeth Donaldson was born on 5 February 1972.
  \item She attended Clear Lake City Elementary School in Houston, Texas, from 1974 to 1977.
  \item Her primary education was at Waimea Heights from 1978 to 1983.
  \item Her secondary schooling was at Taroona High School from 1984 to 1987.
  \item She studied at the University of Tasmania from 1990 to 1994, graduating with a combined Bachelor of Commerce and Bachelor of Laws degree on 27 May 1995.
\end{itemize}

\textbf{Conclusion}

Since 14 May 1988 falls within her secondary schooling period, Mary was attending Taroona High School at that time.

\smallskip
\textbf{Answer:} Mary, Crown Princess of Denmark, was educated at Taroona High School in Hobart, Tasmania, 15 years and 6 months before her marriage to Frederik, Crown Prince of Denmark.
}}
\end{mdframed}

% ── (c) IA-RAG ──────────────────────────────────────────────────────────────
\begin{mdframed}[
  backgroundcolor=greenlight,
  linewidth=0pt,
  innerleftmargin=10pt, innerrightmargin=10pt,
  innertopmargin=3pt, innerbottommargin=3pt,
  skipabove=0pt, skipbelow=6pt,
  roundcorner=4pt,
]
{\small
\textbf{\color{greendark}(c) IA-RAG Answer:}\hfill
\colorbox{rightbg}{\fontsize{8}{10}\selectfont\color{rightgreen}\textbf{$\checkmark$ Correct}}

\medskip
{\color{greendark}
To determine where Mary, Crown Princess of Denmark, was educated 15 years and 6 months before she was married to Frederik, Crown Prince of Denmark, we need to identify the date of her marriage and then go back 15 years and 6 months from that date.

\smallskip
\textbf{Timeline Analysis}

- From Event \#12 (EVT002091): Frederik and Mary married on \textbf{14 May 2004} in Copenhagen Cathedral.\\
- Going back 15 years and 6 months from 14 May 2004 lands us on \textbf{1 November 1988}.

\smallskip
\textbf{Education Verification}

\smallskip
From Event \#9 (EVT002087): Mary Elizabeth Donaldson graduated with a combined Bachelor of Commerce and Bachelor of Laws degree on 27 May 1995.

\smallskip
From Chunk 2: ``She studied at the University of Tasmania from 1990 to 1994\ldots''

\smallskip
Since 1990 is later than 1 November 1988, we need to check earlier educational activities from Chunk 2:

\begin{mdframed}[
  backgroundcolor=greenaccent,
  linewidth=0pt,
  innerleftmargin=8pt, innerrightmargin=8pt,
  innertopmargin=5pt, innerbottommargin=5pt,
  skipabove=2pt, skipbelow=2pt,
  roundcorner=3pt,
]
\textit{``Her primary education, from 1978 to 1983, was at Waimea Heights with her secondary schooling (1984--1987) being at Taroona High School, and matriculation (1988--1989) at Hobart College.''}
\end{mdframed}
\textbf{Conclusion}\\
Therefore, 15 years and 6 months before her marriage in 2004, Mary was finishing her matriculation at Hobart College in Tasmania, Australia.\\
\textbf{Answer:} Mary, Crown Princess of Denmark, was finishing her matriculation at Hobart College in Tasmania, Australia, 15 years and 6 months before she was married to Frederik, Crown Prince of Denmark.
}}
\end{mdframed}

\caption{Case study 3}
\label{tab:case_study_complextr_T3}
\end{table*}

%%%%%%%%%%%%%%%% Prompt Template  %%%%%%%%%%%%%%%%%%%%%%%%%%%%%%%%
\section{Prompt Templates}
\label{app:prompt}
% In this appendix, we provide the prompt templates used in IA-RAG, including interval event unit extraction(Figure~\ref{prompt:ieu_prompt}), IEU-level deduplication(Figure~\ref{prompt:ieu_dedup_prompt}), temporal interval tightening(Figure~\ref{prompt:time_tightening_prompt}), temporal reasoning type classification(Figure~\ref{prompt:types}), and answer generation(Figure~\ref{prompt:iarag_generation_prompt}). These prompts are designed to guide LLMs in temporal grounding, logical inference, and temporally coherent retrieval throughout the IA-RAG pipeline.

% In this appendix, we provide the prompt templates used in IA-RAG, including interval event unit extraction, IEU-level deduplication, temporal interval tightening, temporal reasoning type classification, and answer generation. These prompts guide LLMs in temporal grounding, interval-aware reasoning, hierarchical retrieval, and temporally coherent response generation. The corresponding templates are illustrated in Figures~\ref{prompt:ieu_prompt}--\ref{prompt:iarag_generation_prompt}.
This appendix presents the prompt templates used in IA-RAG, including interval event unit extraction, IEU deduplication, temporal interval tightening, temporal reasoning type classification, and answer generation. The corresponding templates are shown in Figures~\ref{prompt:types}--\ref{prompt:iarag_generation_prompt}.

\clearpage

\begin{figure*}[tp]
  \begin{tcolorbox}[
    breakable,
    enhanced,
    colback=gray!5,
    colframe=gray!40!black,
    boxrule=0.5pt,
    sharp corners=south,
    title=Prompt: Temporal Reasoning Types,
    fontupper=\scriptsize\ttfamily\linespread{1.1}\selectfont
  ]

You are an expert in Temporal Question Answering. Your task is to classify a given question into one of the following temporal reasoning types:\\

\textbf{T1: Time-Anchored (Point or Range-based)}\\
- Definition: The temporal constraint is a FIXED timestamp (Year, Month, or Day).\\
- Includes: "In 2001", "Since 2001", "Before/After May 2001".\\
- Rule: NO other events or numerical durations (e.g., "5 years") are involved.\\
- Example: "Which position did Brian Smith hold after May 2001?"\\

\textbf{T2: Explicit Interval}\\
- Definition: The query specifies a fixed START and END point.\\
- Keywords: "Between X and Y", "From X to Y".\\
- Example: Which team did Attaphol Buspakom play for between May 1989 and Oct 1990?"\\

\textbf{T3: Numerical Offset from Timestamp (Time-to-Time)}\\
- Definition: A specific duration (X years, Y months) applied to a FIXED timestamp.\\
- Logic: [Fixed Date] +/- [Duration].\\
- Rule: Does NOT reference any specific event as the base.\\
- Example: "Who did A work for 8 years before April 1926?"\\
eg. "Which employer did Gustav Ludwig Hertz work for 8 years and 2 months before April 1926?",\\
eg. Which political party did Jozo Radoš belong to 13 years and 2 months before October 2017?\\
eg. Which political party did Marcel Dassault belong to 39 years and 1 months after February 1946?\\

\textbf{T4: Event-to-Event Relation (Purely Relative)}\\
- Definition: One event's time is defined relative to another event WITHOUT numerical offsets.\\
- Keywords: "Before he joined X", "After her term at Y".\\
- Rule: NO numbers/durations (years/months) allowed.\\
- Example: "Which party did Jüri Adams belong to after he was a member of Isamaa?"\\
eg.Where was Beate Meinl-Reisinger educated before he/she was the member of NEOS – The New Austria?",\\
eg. Which political party did Jüri Adams belong to after he/she was the member of Isamaa?\\
eg. Which employers did Marian Salzman work for when he/she held the position of chief strategy officer?\\

\textbf{T5: Multi-hop Reasoning (Event + Offset/Multi-step)}\\
- Definition: Complex reasoning combining events with durations OR linking multiple events.\\
- Logic: [Event A] +/- [Duration] relative to [Event B].\\
- Rule: If you see a "Duration" (e.g., 2 years) AND an "Event" (e.g., "worked at X"), it MUST be T5.\\
- Example: "Which employer did A work for 3 years after he worked for University B?"\\
eg. Which employer did Manuel García Velarde work for 3 years and 11 months after he/she worked for Université libre de Bruxelles?",\\
eg. "Which employer did Selig Brodetsky work for 44 years and 6 months after he/she studied at Central Foundation Boys' School?"\\
eg. Where was Jessica Stegrud educated 29 years before he/she held the position of member of the Swedish Riksdag?\\
Output JSON schema:
\begin{verbatim}
{{
  "type": "T1" | "T2" | "T3" | "T4" | "T5"
}}
\end{verbatim}
Instructions:\\
- Return ONLY valid JSON\\
- Do NOT explain your reasoning\\
- Do NOT infer events or time intervals\\
- Only judge relevance of relations\\

\textbf{Question}:
\{question\}
    \end{tcolorbox}
    \caption{Prompt template used for temporal reasoning type classification.}
  \label{prompt:types}
\end{figure*}

%%%%%%%%%%%%%%%% 提取ieu %%%%%%%%%%%%%%

\begin{figure*}[!ht]
  \begin{tcolorbox}[
    breakable,                  % 允许自动跨页
    colback=gray!5,
    colframe=gray!40!black,
    boxrule=0.5pt,
    sharp corners=south,
    title=Prompt specification for interval event unit extraction,
    % fontupper=\small\ttfamily\linespread{1.1}\selectfont
    fontupper=\scriptsize\ttfamily\linespread{1.1}\selectfont
  ]

\textbf{Role.}
You are an information-extraction assistant for building \emph{dynamic knowledge graphs}.
Your task is to extract \textbf{every Interval Event Unit (IEU)} that can be directly retrieved
from the given text.

An IEU is a \textbf{complete factual statement} (one clause or several tightly linked clauses) that:
(i) occurs at a \textbf{specific time point or interval};
(ii) has \textbf{high information content}; and
(iii) contains an \textbf{explicit subject with no pronouns}.

\vspace{0.5em}
\textbf{1. Temporal Reasoning Protocol}

\emph{1.1 Reference-Time Stack.}
Maintain a reference-time stack by pushing every explicit absolute date
(Year / Year--Month / Year--Month--Day).
The top of the stack serves as the default anchor for resolving relative time expressions.

\emph{1.2 Priority Parsing (apply before other reasoning).}
Prioritize the following patterns:
(i) bracketed dates: (YYYY), (YYYY--MM), (YYYY--MM--DD);
(ii) hyphenated intervals: 1995--1998, 2003--05;
(iii) parenthetical intervals: (1990--1995), (from 2000 to 2005);
(iv) sentence-ending standalone years (e.g., ``established in 1995'').

\emph{1.3 Relative and Vague Time Resolution.}
Resolve ambiguous time expressions using the following precedence:
(1) immediately preceding dated event;
(2) earlier explicit date in the same sentence or paragraph;
(3) document metadata or header timeline;
(4) paragraph-level inheritance from neighboring sentences.

\vspace{0.5em}
\textbf{Time Interval Conversion (YYYY--MM--DD Format).}
All time expressions must be normalized into a closed interval
$[t_{\text{start}}, t_{\text{end}}]$.

\begin{itemize}[leftmargin=1.5em]
  \item \textbf{Exact day (YYYY--MM--DD):}
  $t_{\text{start}} = t_{\text{end}}$, \texttt{FuzzyFlag=false}.
  \item \textbf{Exact month (YYYY--MM):}
  first day to last day of the month, \texttt{FuzzyFlag=false}.
  \item \textbf{Exact year (YYYY):}
  YYYY-01-01 to YYYY-12-31; \texttt{FuzzyFlag=true} for point events
  (e.g., appointment), \texttt{false} for states or durations.
  \item \textbf{Explicit interval (e.g., YYYY--YYYY):}
  expanded to full granularity, \texttt{FuzzyFlag=false}.
  \item \textbf{``Until X'':}
  unknown start, end at X, \texttt{FuzzyFlag=true}.
  \item \textbf{Relative dates (e.g., ``two days later''):}
  resolved via calendar arithmetic, \texttt{FuzzyFlag=false}.
\end{itemize}

\emph{Fuzzy Time Expressions.}
Expressions such as ``early YYYY'', ``mid YYYY'', ``late YYYY'', or ``around YYYY''
must always be marked with \texttt{FuzzyFlag=true} and mapped to their canonical
sub-intervals.

\vspace{0.5em}
\textbf{2. Information-Value Filter} Each candidate sentence is scored from 0 to 4, receiving one point for each satisfied criterion:
\begin{itemize}[leftmargin=1.5em]
  \item \textbf{Specific actor:}
  contains a named entity or definite noun phrase uniquely identifying the subject.
  \item \textbf{Action or change:}
  describes an action, role change, appointment, or bounded continuation.
  \item \textbf{Result or magnitude:}
  includes quantitative detail, outcome, or measurable consequence.
  \item \textbf{Temporal anchoring:}
  time is resolvable to at least month-level precision or bounded by clear intervals.
\end{itemize}

\emph{Retention Rule.}
Keep the sentence if the score $\geq 1$, or if it matches explicit role-duration patterns
(e.g., ``served as \ldots from \ldots to \ldots''),
contains clear temporal markers, or describes a continued state with explicit boundaries.
If temporal anchoring is missing, paragraph-level inheritance must be attempted;
otherwise, discard the sentence.

\textbf{3. Mandatory Subject Rules} Every extracted IEU must contain a clear, explicit subject.
Use the most complete entity form available (full names, official titles, full organization names).
\textbf{No pronouns} (e.g., ``he'', ``she'', ``they'', ``it'') are permitted.
Scan the entire document, including titles, to recover complete entity names.

\textbf{4. Sentence Formation Guidelines} The \texttt{sentence} field must be self-contained, preserve the original time expression,
and contain no pronouns.
The optional \texttt{context} field may include up to 80 tokens of supplementary information
drawn only from the same text chunk.
Never merge information across different time expressions;
sentences containing multiple distinct times must be split into separate IEUs.

\textbf{5. Output Format} Return \textbf{ONLY} a valid JSON object.
No explanations, no additional text.\textbf{JSON Schema.}
\begin{quote}
\texttt{
\{
  "events": [
    \{
      "event\_id": "E1",
      "sentence": "<explicit-subject factual sentence>",
      "context": "<optional, $\leq$ 80 tokens>",
      "t\_start": "YYYY-MM-DD | unknown",
      "t\_end": "YYYY-MM-DD | unknown",
      "FuzzyFlag": true | false
    \}
  ]
\}
}
\end{quote}
If no valid IEU can be extracted, or if any subject cannot be fully resolved
(e.g., remaining pronouns), return:
\begin{quote}
\texttt{\{ "events": [] \}}
\end{quote}
\textbf{Example.}\emph{Input Text.}  
Professor Lin began directing the Quantum Materials Laboratory in early 2018.
From 2019 to 2021, the laboratory developed three prototype superconducting chips.
Two months after completing the third prototype, Professor Lin signed a collaboration
agreement with Tsinghua University. Later that year, the joint team published
preliminary findings, though the exact month was not specified. \emph{Expected Output (JSON).}
\begin{quote}
\texttt{
\{
  "events": [
    \{
      "event\_id": "E1",
      "sentence": "Professor Lin began directing the Quantum Materials Laboratory in early 2018.",
      "context": "",
      "t\_start": "2018-01-01",
      "t\_end": "2018-04-30",
      "FuzzyFlag": true
    \},
    \{
      "event\_id": "E2",
      "sentence": "The Quantum Materials Laboratory developed three prototype superconducting chips from 2019 to 2021.",
      "context": "",
      "t\_start": "2019-01-01",
      "t\_end": "2021-12-31",
      "FuzzyFlag": false
    \},
    \{
      "event\_id": "E3",
      "sentence": "Professor Lin signed a collaboration agreement with Tsinghua University two months after completing the third prototype.",
      "context": "The third prototype was completed within 2021.",
      "t\_start": "2022-02-01",
      "t\_end": "2022-02-28",
      "FuzzyFlag": true
    \},
    \{
      "event\_id": "E4",
      "sentence": "The joint team published preliminary findings later that year.",
      "context": "The publication occurred after the collaboration agreement.",
      "t\_start": "2022-09-01",
      "t\_end": "2022-12-31",
      "FuzzyFlag": true
    \}
  ]
\}
}
\end{quote}
\textbf{Notice.}
\begin{itemize}
  \item If you are uncertain about extracting any IEU, or cannot fully resolve
  all pronouns to explicit entities, output \texttt{\{ "events": [] \}}.
  \item Do \textbf{not} reveal internal scoring, heuristics, or reasoning steps.
  \item Sentences should remain maximally complete and precise to support
  direct downstream retrieval as event evidence.
\end{itemize}
\textbf{Real Data Interface.} Text: \{input\_text\}
% \begin{quote}
% \texttt{
% Text: \{input\_text\}
% }
% \end{quote}
  \end{tcolorbox}
\caption{Prompt template used for interval event unit extraction.}
  \label{prompt:ieu_prompt}
\end{figure*}

\begin{figure*}[tp]
  \begin{tcolorbox}[
    breakable,                  % 允许自动跨页
    colback=gray!5,
    colframe=gray!40!black,
    boxrule=0.5pt,
    sharp corners=south,
    title=Prompt: Strict IEU-level Deduplication,
    fontupper=\scriptsize\ttfamily\linespread{1.1}\selectfont
    % fontupper=\small\ttfamily\linespread{1.1}\selectfont
  ]
Role. You are a Strict Data Auditor for IEU (Interval Event Unit) deduplication.
This task is ONLY about filtering redundant records.
This is NOT summarization, NOT aggregation, and NOT timeline construction. \\

The ``Atomic IEU'' Rule (CRITICAL).
A valid IEU must represent exactly one atomic fact:\\
- One subject\\
- One specific role or action\\
- One single continuous time interval or time point\\[1em]

Task. Analyze \texttt{\{len(events)\}} IEUs and determine whether any subset refers to the
\textbf{exact same real-world occurrence}.\\
- If events are distinct (even if related), \textbf{do not merge}. \\
- Only merge \textbf{redundant mentions of the same atomic fact}.\\ \\

Strict Prohibitions (Do \emph{Not} Merge If):
 - Organization mismatch: different institutions imply distinct events (e.g., ``Yale University'' vs.\ ``Harvard University'').\\
- Sequential chaining: do not merge consecutive career stages. \\
- Role or title change: e.g., ``Assistant Professor'' vs.\ ``Dean''.\\
- Information loss: merging would obscure a change in location or institution. \\
- Temporal chaining: do not construct a career summary.\\
- Multiple time references: a sentence must not mention more than one time span.\\
- Time gaps: if there is a temporal gap, events must remain separate.\\

Allowed Merge (Only If): \\
The IEUs are redundant mentions of the same single occurrence
at the same organization and during the same continuous period.\\
- Example: ``A played for AC Milan in 2010'' + ``A contract at AC Milan from 2010 to 2012'' $\rightarrow$ merge into a single event spanning 2010--2012. \\

Merging Rules for Sentence and Time:\\
- Sentence: must be canonical, atomic, and contain no ``and''. \\
- Precision first: prefer Day $>$ Month $>$ Year.\\
- No expansion: the merged interval must not exceed the real duration of a single appointment or stay. \\

Notes.\\
- \texttt{merged\_event\_ids} must include \emph{only} the merged IEUs.\\
-  All other IEUs are implicitly preserved without modification.\\

Input IEUs:
\begin{quote}
\texttt{\{events\_list\_str\}}
\end{quote}

Output Format (Strict).
Return \textbf{only} a valid JSON object. Do not include explanations.

If no events should be merged:
\begin{verbatim}
{
  "if_merge": false
}
\end{verbatim}

If one mergeable subset is found:
\begin{verbatim}
{
  "if_merge": true,
  "merged_event_ids": ["event_global_id1", "event_global_id2"],
  "sentence": "Single atomic sentence with no conjunctions.",
  "context": "Integrated context from all source IEUs.",
  "t_start": "YYYY-MM-DD",
  "t_end": "YYYY-MM-DD",
  "FuzzyFlag": true
}
\end{verbatim}
  \end{tcolorbox}
  \caption{Prompt template used for strict IEU-level deduplication and redundancy removal.}
\label{prompt:ieu_dedup_prompt}
\end{figure*}

\begin{figure*}[tp]
  \begin{tcolorbox}[
    breakable,
    enhanced,
    colback=gray!5,
    colframe=gray!40!black,
    boxrule=0.5pt,
    sharp corners=south,
    title=Prompt: Temporal Interval Tightening and Logical Inference,
    fontupper=\scriptsize\ttfamily\linespread{1.1}\selectfont
  ]

\textbf{Role}\\
You are a \textbf{High-Precision Temporal Auditor}. Your role is to perform
\textbf{Temporal Interval Tightening and Logical Inference} on a set of
semantically linked Interval Event Units (IEUs).\\

\textbf{Task Goal}\\
1. \textbf{Tightening}: Identify IEUs marked as \texttt{FuzzyFlag: true}
   and refine them into precise intervals using neighboring evidence.\\
2. \textbf{Inference}: Reason about \texttt{unknown} or \texttt{fuzzy} dates
   by leveraging the \textbf{causal or sequential logic} within the connected
   component.\\

\textbf{Core Constraints}\\
\begin{itemize}[leftmargin=*, nosep]
  \item \textbf{Absolute Accuracy}: Every update must be anchored in the
        provided text. NO world knowledge.
  \item \textbf{Atomic Preservation}: Do NOT merge, rewrite, or reinterpret
        events. Only update \texttt{t\_start}, \texttt{t\_end}, and
        \texttt{FuzzyFlag}.
  \item \textbf{Directional Logic}: A refinement must narrow the uncertainty,
        never expand it.
\end{itemize}

\textbf{Time Interval Conversion (YYYY-MM-DD Format)}\\
All time expressions must be converted into a precise \texttt{[t\_start,
t\_end]} interval.

\medskip
\begin{tabular}{@{}llllll@{}}
\hline
\textbf{Rule} & \textbf{Expression Type} & \textbf{t\_start}
  & \textbf{t\_end} & \textbf{FuzzyFlag} & \textbf{Notes}\\
\hline
A & Exact Day (YYYY-MM-DD)   & Same Day          & Same Day         & false & \\
B & Exact Month (YYYY-MM)    & 1st of Month      & Last of Month    & false & \\
C & Exact Year (YYYY)        & YYYY-01-01        & YYYY-12-31       & See below & \\
D & Explicit Interval        & Start Boundary    & End Boundary     & false & Full granularity\\
E & ``Until X'' (Duration End) & unknown         & Date X           & true  & Start not explicit\\
G & Relative Date            & Calendar Arith.   & Calendar Arith.  & false & Anchored precisely\\
\hline
\end{tabular}

\medskip
\textbf{FuzzyFlag Specifics:}
\begin{itemize}[leftmargin=*, nosep]
  \item \textbf{Year (C):} \texttt{true} for point events (join, begin,
        appointment); \texttt{false} for states/durations (served during 1999).
  \item \textbf{Fuzzy Expressions (F):} Always \texttt{true}.
  \item \textbf{Other True Conditions:} Time is inferred, approximated,
        expanded by default rules, or based on a fuzzy anchor.
\end{itemize}

\medskip
\textbf{Fuzzy Time \& Interval Table}

\medskip
\begin{tabular}{@{}llll@{}}
\hline
\textbf{Expression} & \textbf{t\_start} & \textbf{t\_end} & \textbf{FuzzyFlag}\\
\hline
Year--year     & YYYY1-01-01   & YYYY2-12-31        & false\\
Month--month   & YYYY1-MM1-01  & YYYY2-MM2-last\_day & false\\
Early YYYY     & YYYY-01-01    & YYYY-04-30         & true\\
Mid YYYY       & YYYY-05-01    & YYYY-08-31         & true\\
Late YYYY      & YYYY-09-01    & YYYY-12-31         & true\\
Around YYYY    & YYYY-01-01    & YYYY-12-31         & true\\
Early YYYY-MM  & YYYY-MM-01    & YYYY-MM-10         & true\\
Mid YYYY-MM    & YYYY-MM-11    & YYYY-MM-20         & true\\
Late YYYY-MM   & YYYY-MM-21    & YYYY-MM-last       & true\\
\hline
\end{tabular}

\medskip
\textbf{Strict Prohibitions (Zero Tolerance)}
\begin{itemize}[leftmargin=*, nosep]
  \item \textbf{NO Hallucination}: Do not use world knowledge.
  \item \textbf{NO Merging}: Do not combine two sentences into one.
  \item \textbf{NO Speculation}: If logic does not strictly dictate a date,
        keep \texttt{FuzzyFlag: true}.
\end{itemize}

\medskip
\textbf{Input Data (Connected Component)}\\
\{events\_list\_str\}

\medskip
\textbf{Output Instructions (JSON ONLY)}\\
If no refinement is possible:
\begin{verbatim}
{ "if_update": false }
\end{verbatim}
If one or more IEUs can be updated:
\begin{verbatim}
{
  "if_update": true,
  "updates_ieus": [
    {
      "event_global_id": "EVT000123",
      "updated_t_start": "YYYY-MM-DD",
      "updated_t_end": "YYYY-MM-DD",
      "updated_FuzzyFlag": false,
      "reasoning": "Evidence-based justification referencing other IEUs."
    }
  ]
}
\end{verbatim}

  \end{tcolorbox}
  \caption{Prompt template used for temporal interval tightening and logical temporal inference.}
\label{prompt:time_tightening_prompt}
\end{figure*}

%%%%%%%%%%% IA-RAG Answer Generation %%%%%%%%%%
\begin{figure*}[tp]
  \begin{tcolorbox}[
    breakable,
    enhanced,
    colback=gray!5,
    colframe=gray!40!black,
    boxrule=0.5pt,
    sharp corners=south,
    title=Prompt: IA-RAG Answer Generation,
    % fontupper=\footnotesize\ttfamily\linespread{1.1}\selectfont
    fontupper=\scriptsize\ttfamily\linespread{1.1}\selectfont
  ]

You are an AI assistant that answers questions based on both temporal event
sequences and relevant text chunks.

\textbf{TASK}\\
Answer the question using both the provided events and text chunks. The events
show temporal relationships, while the text chunks provide additional context
and details.\\

\textbf{TEMPORAL RELEVANCE GUIDELINE}\\
\textbf{Prioritize information from time periods that align with the question's
temporal scope.} If a question asks about a specific time period, focus on
events and information that occurred within or near that timeframe.\\

\textbf{ANALYSIS APPROACH}
\begin{enumerate}[leftmargin=*, nosep, topsep=2pt]
  \item \textbf{Query Semantics Understanding}: Analyze the semantic intent:
    \begin{itemize}[leftmargin=*, nosep]
      \item \textbf{Existential queries}: What exists/happens at a specific time
      \item \textbf{Continuity queries}: Ongoing states, processes, or relationships
      \item \textbf{Boundary queries}: Beginnings, endings, or transitions
      \item \textbf{Aggregate queries}: Synthesis across multiple time points
    \end{itemize}

  \item \textbf{Temporal Scope Identification}: Identify exact time constraints:
    \begin{itemize}[leftmargin=*, nosep]
      \item \textbf{Subject Consistency Check}: Verify the event's subject matches the question's focus
      \item Extract specific dates, time periods, or temporal references
      \item Mark temporal qualifiers (e.g., ``during'', ``between'', ``before'', ``after'')
    \end{itemize}

  \item \textbf{Evidence Time Filtering}: Before using any event or chunk:
    \begin{itemize}[leftmargin=*, nosep]
      \item \textbf{Assess temporal relevance}: Evaluate whether evidence falls within or near the question's time scope
      \item \textbf{Consider contextual value}: Include nearby time periods if essential background
      \item \textbf{Prioritize temporal proximity}: Give preference to evidence closer to the target time
    \end{itemize}

  \item \textbf{Event Analysis}: Review the temporal sequence of events:
    \begin{itemize}[leftmargin=*, nosep]
      \item \textbf{Temporal Ordering}: Analyze chronological relationships
      \item \textbf{State Persistence and Change}: Persistent states / Instantaneous events / Processes / Transitions
      \item \textbf{Office/Role Incumbency Reasoning}: Treat holding an office as a persistent state;
            if no explicit end event is found before the query time, assume the holder remains in position
    \end{itemize}

  \item \textbf{Entity-Event Relationships}: Analyze connections:
    \begin{itemize}[leftmargin=*, nosep]
      \item \textbf{Agent}: Who performs or causes an action
      \item \textbf{Patient}: Who/what is affected
      \item \textbf{Locative}: Where something happens
      \item \textbf{Attributive}: Properties or characteristics at specific times
    \end{itemize}

  \item \textbf{Chunk Analysis}: Extract relevant information from text chunks

  \item \textbf{Cross-Reference}: Connect information between events and chunks
\end{enumerate}

\textbf{RESPONSE GUIDELINES}
\begin{itemize}[leftmargin=*, nosep, topsep=2pt]
  \item \textbf{TEMPORAL RELEVANCE ASSESSMENT}: Assess whether evidence timestamps are relevant
  \item Reference specific events (e.g., ``Event \#3'') or chunks when supporting your answer
  \item Apply temporal logic: if an event occurred at T1 and no change is mentioned by T2, assume continuity
  \item \textbf{If limited information exists}: state temporal limitations, then infer based on:
    (1) Career Continuity, (2) Role Persistence, (3) Institutional Affiliation --- label inferences clearly
  \item Start with the direct answer followed by justification citing key events
  \item If based on inference, state: ``Based on inference from surrounding evidence: [answer]''
\end{itemize}

\medskip
\textbf{QUESTION}\\
\{question\}

\medskip
\textbf{EVENTS}\\
\{events\_data\}

\medskip
\textbf{TEXT CHUNKS}\\
\{chunks\_data\}

\medskip
If there is any conflict between EVENTS and TEXT CHUNKS, \textbf{TEXT CHUNKS
take precedence} and must be trusted as the authoritative source.

\medskip
\textbf{ANSWER}

  \end{tcolorbox}
  \caption{Prompt template used for IA-RAG answer generation based on retrieved temporal evidence.}
\label{prompt:iarag_generation_prompt}
\end{figure*}

\end{document}